\documentclass[journal]{IEEEtran}
\usepackage{times}
\usepackage{soul}
\usepackage{url}
\usepackage[utf8]{inputenc}
\usepackage[small]{caption}
\usepackage{graphicx}
\usepackage{amsmath}
\usepackage{amsthm}
\newtheorem{proposition}{Proposition}
\usepackage{booktabs}
\usepackage{epstopdf}
\usepackage{algorithm}
\usepackage{algorithmicx}
\usepackage{algpseudocode}
\usepackage{subfigure}
\usepackage{array}
\usepackage{multirow}
\usepackage{xcolor}
\usepackage{amssymb}
\usepackage{cite}
\usepackage{ulem}
\usepackage{lineno}

\usepackage{tikz}
\usetikzlibrary{calc}
\usetikzlibrary{arrows.meta}
\ifCLASSINFOpdf
\else
\fi
\begin{document}
%
\title{Density-aware Haze Image Synthesis by Self-Supervised Content-Style Disentanglement}
%
%
%
%

\author{Chi Zhang$^*$,~\IEEEmembership{Member,~IEEE,}
        Zihang Lin$^*$, Liheng Xu, Zongliang Li \\ Wei Tang,~\IEEEmembership{Member,~IEEE,} Yuehu Liu,~\IEEEmembership{Member,~IEEE,} Gaofeng Meng,~\IEEEmembership{Senior Member,~IEEE,}\\
         Le Wang,~\IEEEmembership{Senior Member,~IEEE,} and Li Li,~\IEEEmembership{Fellow,~IEEE}
\IEEEcompsocitemizethanks{
\IEEEcompsocthanksitem  $^*$Equal contribution.
\IEEEcompsocthanksitem  This work was supported in part by the National Key Research and Development Project of New Generation Artificial Intelligence of China under Grant 2018AAA0102504, and in part by the National Natural Science Foundation of China under Grant 61973245. Yuehu Liu is the corresponding author. E-mail: \texttt{liuyh@xjtu.edu.cn}.
\IEEEcompsocthanksitem Chi Zhang, Zihang Lin, Liheng Xu, Zongliang Li, Yuehu Liu and Le Wang are with the Institute of Artificial Intelligence and Robotics, Xi'an Jiaotong University, Xi'an, China. 
\IEEEcompsocthanksitem Wei Tang is with the Department of Computer Science, University of Illinois at Chicago, Chicago, IL 60607, USA.
\IEEEcompsocthanksitem Gaofeng Meng is with the National Laboratory of Pattern Recognition, Institute of Automation, Chinese Academy of Sciences, Beijing 100190, China.
\IEEEcompsocthanksitem Li Li is with the Department of Automation, BNRist, Tsinghua University, Beijing 100084, China.
}
\thanks{Manuscript is received on Jun 30, 2021, revised on Sep 2, 2021 and accepted on Nov 16, 2021.}}

%
%

\markboth{IEEE Transactions on Circuits and Systems for Video Technology,~Vol.~, No.~, April~2021}%
{Shell \MakeLowercase{\textit{et al.}}: Bare Demo of IEEEtran.cls for Computer Society Journals}
%


\makeatletter
\def\ps@IEEEtitlepagestyle{
	\def\@oddfoot{\mycopyrightnotice}
	\def\@evenfoot{}
}
\def\mycopyrightnotice{
	{\footnotesize
		\begin{minipage}{\textwidth}
			\centering
			Copyright~\copyright~2021 IEEE. Personal use of this material is permitted. However, permission to use this  \\ 
			material for any other purposes must be obtained from the IEEE by sending a request to \texttt{pubs-permissions@ieee.org}.
		\end{minipage}
	}
}


\IEEEtitleabstractindextext{%
\begin{abstract}
The key procedure of haze image synthesis with adversarial training lies in the disentanglement of the feature involved only in haze synthesis, i.e., \textit{the style feature}, from the feature representing the invariant semantic content, i.e., \textit{the content feature}. Previous methods introduced a binary classifier to constrain the domain membership from being distinguished through the learned content feature during the training stage, thereby the style information is separated from the content feature. However, we find that these methods cannot achieve complete content-style disentanglement. The entanglement of the flawed style feature with content information inevitably leads to the inferior rendering of haze images. To address this issue, we propose a self-supervised style regression model with stochastic linear interpolation that can suppress the content information in the style feature. Ablative experiments demonstrate the disentangling completeness and its superiority in density-aware haze image synthesis. Moreover, the synthesized haze data are applied to test the generalization ability of vehicle detectors. Further study on the relation between haze density and detection performance shows that haze has an obvious impact on the generalization ability of vehicle detectors and that the degree of performance degradation is linearly correlated to the haze density, which in turn validates the effectiveness of the proposed method.
\end{abstract}

}

\maketitle

\IEEEdisplaynontitleabstractindextext

%
\IEEEpeerreviewmaketitle

\section{Introduction}\label{sec:introduction}

\begin{figure}[!tbh]
	\centering
	\includegraphics[width=\linewidth]{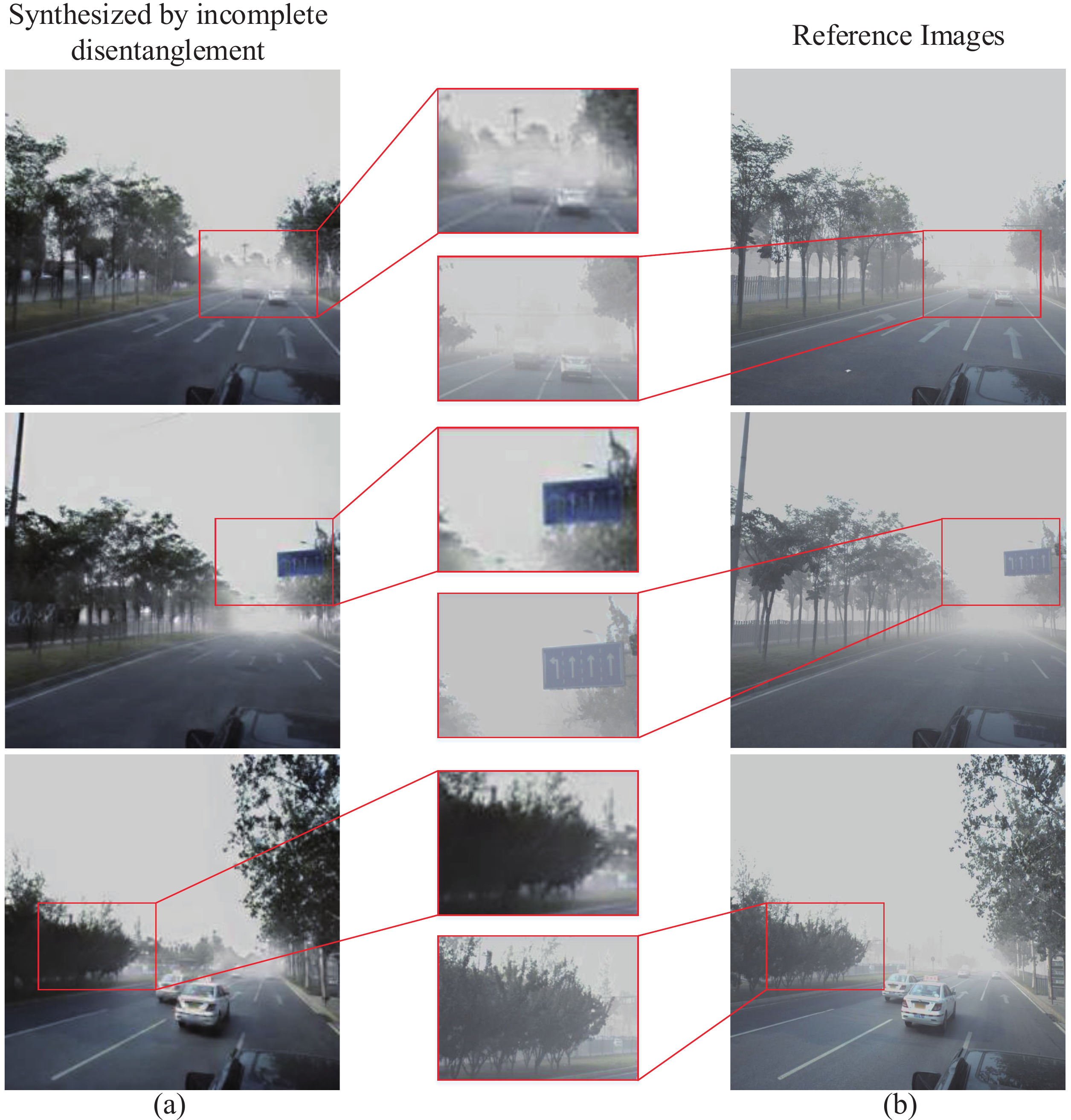}
	\setlength{\abovecaptionskip}{-5pt}
	\setlength{\belowcaptionskip}{-5pt}
	\caption{Examples of unreal haze images caused by incomplete disentanglement. (a) presents haze images synthesized without complete content-style disentanglement. As the image in the top row shows, the rendered haze appears abruptly (similar to a puff of smoke) on the road, while the haze in the corresponding reference images (b) is evenly distributed according to the depth and semantic content. In addition, objects such as the signs and trees in the middle and bottom images, respectively, are excluded from the haze. This uneven distribution makes the synthesized images look factitious.}
	\label{Fig2}
\end{figure}

\IEEEPARstart{F}{og} and haze cause serious issues with regard to the driving safety of vehicles, especially for autonomous vehicles equipped with optical cameras and LiDAR sensors. The image data collected by these sensors degrade heavily due to the densely distributed atmospheric suspended particles. However, since foggy and hazy weather is temporary and difficult to forecast, there exist limited samples of diverse concentration levels under these scenarios. As a result, there is a lack of sufficient data for autonomous perception, especially when testing its generalization in terms of hazy weather.


To acquire sufficient haze images for testing, a common practice is to synthesize them from clear images and the corresponding depth maps under the haze image formation model. Many sophisticated handcrafted methods such as those of \cite{sun2015algorithm}\cite{sarker2019simulation} exploit or calculate depth maps to estimate the atmosphere transmittance (scattering ability of suspended particles) in the environment. In this case, the quality of the synthesized images depends on the accuracy of the depth prediction to some extent. However, compared with color images, depth images acquired by professional equipment are often noisy and discontinuous \cite{mallick2014characterizations}\cite{sweeney2019supervised}. On the other hand, the process of depth prediction will result in extra errors. These inherent defects make it difficult for the traditional algorithms mentioned above to break through the bottleneck of haze image synthesis. As a novel solution to image synthesis problems, generative adversarial networks \cite{Goodfellow2014Generative} have seen extensive applications in various tasks, including image translation \cite{Isola2016Image,Zhu2017Unpaired,wenzel2018modular,huang2018multimodal,Liu2017Unsupervised}, image inpainting \cite{xu2020e2i, ge2020occluded, pathak2016context}, and face recognition \cite{du2019age, zhang2021joint, xia2021local}. Inspired by previous works \cite{jaw2020desnowgan, zhang2019image, ding2021rain, zhu2020see} that explore how to transform the weather conditions between image domains using generative adversarial networks, the generation of haze images can also be carried out based on the idea of image translation, in which a style transformation between the haze domain and the haze-free domain is performed.

Theoretically, it is impossible to obtain the paired training images required for supervised image-translation algorithms because a particular scene is either hazy or clear at a particular moment. Instead, this problem is posed as unsupervised image translation, which does not require a one-to-one correspondence. For a pair of images, i.e., a haze image and a clear image, captured in the same scene, they are assumed to be disentangled into the feature involved only in haze synthesis, i.e., \textit{the style feature}, and the feature representing the invariant semantic content, i.e., \textit{the content feature}. Typically, the existing image-translation methods decouple image content and style features and introduce strict constraints to guarantee the absolute separateness of style information from the content feature. Nevertheless, potential content information remaining in the style feature is ignored, which degrades the synthesized image in terms of photorealism. As shown in Fig.~\ref{Fig2}, due to this incomplete disentanglement, the rendered haze simply appears abruptly, similar to a puff of smoke, covering the end of the road, which looks factitious. Moreover, haze images simply generated by such image translation lack diversity since the network is unaware of the haze density. 

In this paper, we address the synthesis issues related to the aforementioned disentangling completeness and density-aware diversity. Benefiting from disentangled representation learning, we extract the content feature of a haze-free image and then combine it with the style feature of a haze image to synthesize the corresponding haze image. Different from \cite{lee2018diverse}\cite{denton2017unsupervised}, we not only introduce the content discriminator to supervise the updating of the content encoder but also apply the regularization of density-aware style feature regression to suppress potential content information in the extracted haze feature. Taking advantage of the density variability of haze and the immutability of content, this process is made self-supervised via randomly linear interpolation of the style feature. Our empirical studies on feature disentanglement prove that we can successfully disentangle the style feature and the content feature. After mapping the style feature to the linear space, we successfully synthesize multi-density haze images by manipulating the style feature. Finally, to demonstrate that the haze images synthesized by our method provide sufficient and effective difficult scenario data support for the scenario test of the autonomous driving perception module, we set up an experiment involving a worst perception scenario (WPS) search with controllable haze density for controlling the difficulty of the scenario. The results indicate that haze density is highly correlated with scene difficulty, which further clarifies the necessity of testing self-driving cars under multi-density haze weather, and that our framework can address the lack of haze data.

The contribution of this paper is threefold:

1. We propose self-supervised style regression to suppress content information in the style feature.

2. By mapping the style feature to the linear space, we transform the domain into a continuous space and present a density-aware disentangling framework of unsupervised image-to-image translation to produce multi-density haze images.

3. Based on our unsupervised framework, which enables the synthesis of diversity-controllable haze images, we quantitatively prove that traffic scenes become worse in haze weather by carrying out an online search for the worst haze scenario.

\begin{figure*}
	\centering
	\includegraphics[width=\linewidth]{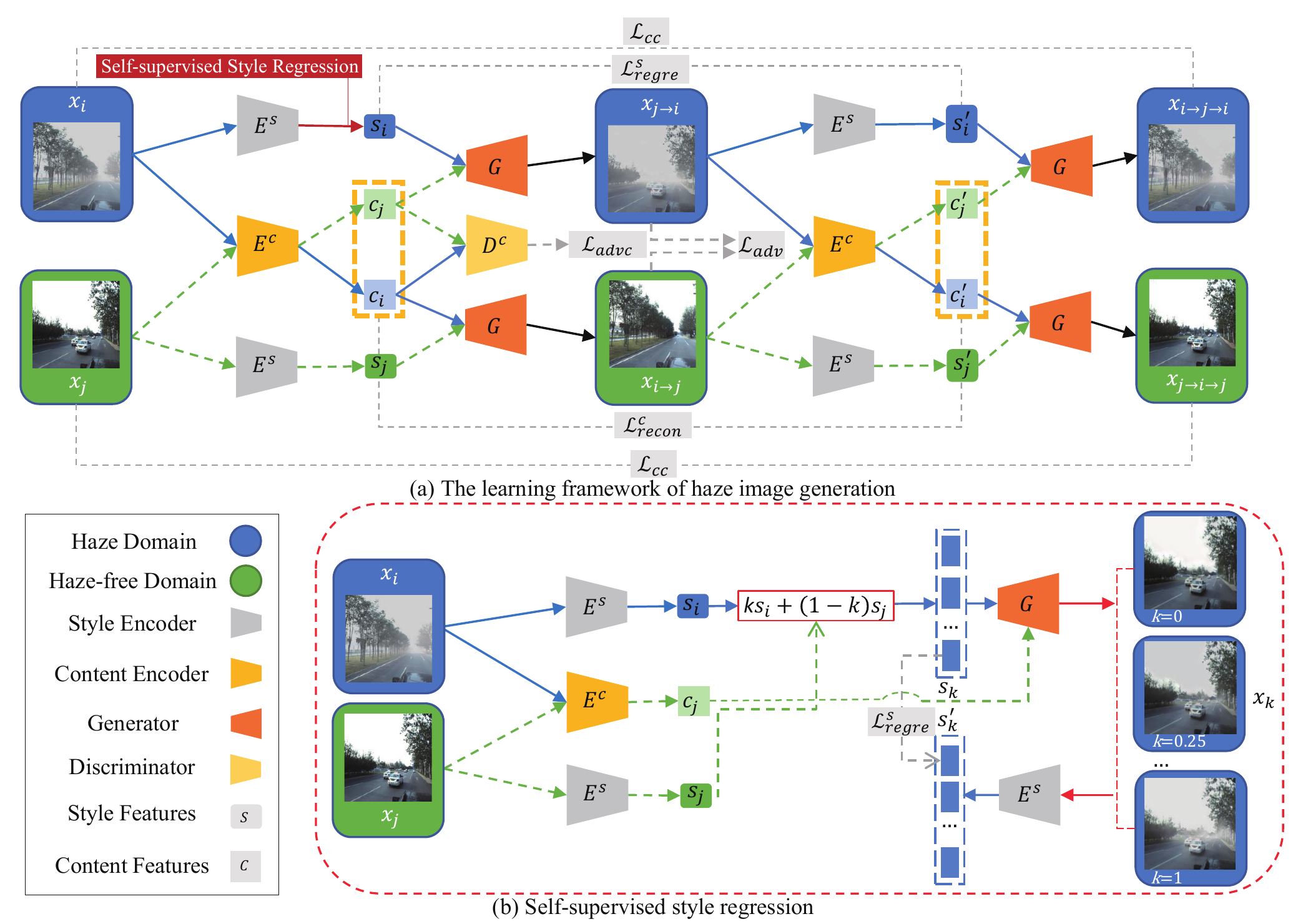}
	\setlength{\abovecaptionskip}{-3pt}
	\setlength{\belowcaptionskip}{-2pt}
	\caption{The overall framework of our multi-density haze image synthesis model. (a){\color{black}{ shows the learning process of haze image synthesis. Following the classic image translation scheme with cycle consistency,  an image from one domain is translated into the other domain and then mapped back. For example, a haze image $x_i$ is first translated into a haze-free image $x_{i \rightarrow j}$ by passing its content feature $c_i$ (extracted by the content encoder $E^c$) to the generator $G$ with the style feature $s_j$ (extracted by the style encoder $E^s$) from the haze-free image $x_j$. Afterwards, the synthesized $x_{i \rightarrow j}$ is mapped back to $x_{i \rightarrow j \rightarrow i}$ in the haze domain in a similar way. With the cross-cycle consistency loss $\mathcal{L}_{cc}$, the adversarial loss $\mathcal{L}_{adv}$ and the content reconstruction loss $\mathcal{L}_{recon}^x$, the mapping between haze domain and haze-free domain is learned with unpaired data. Moreover, content discriminator $D_c$ and $\mathcal{L}_{advc}$ are incorporated to suppress potential style information in the extracted content feature \cite{lee2018diverse}. With the proposed self-supervised style regression, the framework is capable of synthesizing haze images with different density level by a single input haze-free image. (b) shows the process of self-supervised style regression. At each iteration, $s_i$ and $s_j$ are interpolated with the randomly selected $k$ to obtain $s_k$, representing a middle density of haze. Then we use $s_k$ as a pseudo label to make the updating of the style encoder a self-supervised process. It is first combined with $c_j$ to synthesize an image $x_k$. Afterwards, the style feature of $x_k$, i.e., $ s_k' $, is aligned to $s_k$ by minimizing the style regression loss $\mathcal{L}_{regre}^s$. }}}
	\label{Fig3}
\end{figure*}

\section{Related work}
The method in this paper benefits from traditional haze image synthesis methods and disentangled image representation learning.
Therefore, this section mainly introduces those works related to haze image
synthesis and disentangled representation learning.

The atmospheric scattering model,  describing the formation of haze images, is formulated mathematically as follows\cite{ju2021idbp}:
\begin{equation}
      I(x) = J(x)t(x) + A(1-t(x))
      \label{equ:haze}
\end{equation}
where $I(x)$ represents the haze image collected by the camera and $J(x)$
denotes the real reflected light on the surface of objects (normal image).
$t(x)$ is the medium transmittance, and $A$ is the atmospheric light intensity.
$x$ represents the pixel coordinate. Given the image depth map $d(x)$ and the atmospheric scattering coefficient $\beta$, the transmittance $t(x)$ can be calculated as $t(x) = e^{-\beta d(x)}$.

Therefore, the key to generating a haze image is to estimate the depth image
of a single haze-free image and then forecast the atmospheric transmittance
(scattering ability of suspended particles) in the imaging environment. Zhuo et
al. estimated the defocus map of a single image by propagating the amount of
defocus blur at the edge of the image to the entire image and estimated the
depth of a single image \cite{zhuo2011defocus}. However, since it is
impossible to accurately determine whether the blur at the edge of the image is
caused by the defocus of the device or a device imaging problem, this
method causes a large error when reckoning the amount of defocus blur. In 2015, Sun
\textit{et al}. simulated the haze under different levels of visibility based on a single image's defocus map \cite{sun2015algorithm}.
Although their method is in accordance with the principle of atmospheric scattering, the oversimplification of smog cannot capture the
randomness and variability of real smog seen in nature due to the simplicity of the
calculation process. 
With the same purpose of rendering
realistic haze, Guo  proposed a new rendering method based on
transmission map estimation \cite{guo2014foggy}. Then, they took advantage of the
Markov random field model and the bilateral filter to estimate the transmission
map and realistically rendered virtual foggy scenes with the synthesized perlin noise image and the transmission map according to the atmospheric scattering model. Nevertheless, compared to ordinary RGB images, depth images are more difficult to acquire. Moreover, the quality of the synthesized image depends on the quality of the depth image, which leads to the existence of obviously unreal parts in the synthesized image.

On the other hand, thanks to the development of deep neural networks and
generative adversarial networks (GANs), the data-driven haze image synthesis
method can be used to mine the mapping relationship between haze-free traffic scene
images and haze images through adversarial learning. In fact, GANs have been widely used to tackle such problems concerned with the change in weather conditions. To remove rain streaks from images, Zhang  used a generative model and utilized perceptual loss to directly erase rain traces from images \cite{zhang2019image}, while Ding  synthesized corresponding depth maps via a GAN to guide the elimination of rain streaks from rainy light-field images \cite{ding2021rain}. To de-snow an image, Jaw  proposed a two-stage deep network that includes a GAN framework, which is used to refine the resulting image \cite{jaw2020desnowgan}. Considering unclear images collected in a dim environment, a GAN was trained to convert nighttime video to daytime video with background regions preserved \cite{zhu2020see}. To remove haze from images, Qu  applied an extra enhancer on the basis of supervised framework pix2pixHD \cite{wang2018high} to improve the quality of output images \cite{qu2019enhanced}. Engin  presented Cycle-dehaze, which combined cycle-consistency and perceptual losses to synthesize visually better haze-free images with unordered training data. \cite{engin2018cycle}. Although these studies realized the removal of adversarial weather conditions from images, their methods can essentially be regarded as an image-translation
model that maps the image in one domain to the
corresponding image \cite{Isola2016Image} in another domain. Image-translation
problems can be solved in a supervised or unsupervised manner. The ``pix2pix"
framework proposed by Isola  is trained in a supervised manner, using
the conditional GAN to learn the mapping between the input image domain and the
output image domain \cite{Isola2016Image}. Unsupervised image-translation models
require only two sets of independent images, that is, they learn the mapping
between two image domains. Because training data are easier to obtain,
unsupervised image-translation models are more widely used. The problem of
unsupervised image translation is ill-posed; thus, it requires more constraints
to be solvable. The cycle consistency constraint proposed in CycleGAN is the
most widely used constraint \cite{Zhu2017Unpaired}. In addition to one-to-one
mapping, the developers of MUNIT \cite{huang2018multimodal} and BicycleGAN \cite{zhu2017toward}
proposed multimodal generative models to achieve one-to-many mapping, in which
multiple images of the target image domain corresponding to the input
image can be synthesized. StarGAN \cite{choi2018stargan} and
UDFN \cite{liu2018unified} further implement image mapping between multiple image
domains. The unsupervised image-translation model can directly learn the mapping
between a haze-free image and a haze image to realize haze image
synthesis. Although general image-translation models can realize the
mapping between haze-free images and haze images, the incomplete disentanglement
of content features and haze features often results in content change, which is
undesirable. Compared with the aforementioned methods \cite{zhang2019image, ding2021rain,jaw2020desnowgan,zhu2020see}, which apply the image-translation model to address the transformation between different weather conditions, this paper uses the disentangled content and haze features to ensure the consistency of the content of synthesized images and input images.

Disentangled representation learning involves modeling the factors that cause data
to change and acquiring interpretable variables that can be identified and
disentangled. Mathieu combined variational autoencoders and the conditional GAN \cite{mirza2014conditional} and
disentangled the representation of data into two domains, that is, category-related
and category-independent, in a weakly supervised learning
manner \cite{mathieu2016disentangling}. Recently, researchers proposed
unsupervised disentangled representation learning. By maximizing the mutual
information between hidden variables and data variables, InfoGAN achieves
unsupervised disentangled representation learning \cite{chen2016infogan}.
In contrast, DrNet uses the viewpoint correlation between a video and adversarial loss
to decouple the characterization of a video frame into a time-independent content
part and time-related pose part \cite{denton2017unsupervised}, while DRIT uses the content adversarial loss to disentangle images into domain-invariant and domain-specific representations \cite{lee2018diverse}. However, the content confrontation loss can only suppress the style information in content features and cannot guarantee that there is no content information in the style features. In this paper, a self-supervised style regression is introduced to reduce the content information in the style features.


\section{Methods}

The task of synthesizing realistic haze images can be formulated as an
adversarial learning process, as shown in Fig.~\ref{Fig3}(a)~\cite{8243966}. Our
goal is to learn the mapping functions between image sets $X_i$ from the haze domain
and $X_j$ from the haze-free domain. Typically, a generator $G$ is utilized to
realize the transfer between the two domains, and a discriminator $D$ is utilized to
distinguish between real images and translated images. Similar to the idea of
intrinsic image decomposition, where an image could be decomposed into the
underlying structure and the texture style, traffic images are assumed to be
decomposed into two parts: semantic contents and texture appearance~\cite{wang2016generative, 8243966}. Based on this notion, 
we propose the following proposition regarding image feature disentanglement:

\begin{proposition}
For an unpaired set that includes haze image $x_i$ and haze-free image $x_j$ captured in the same scenario, the feature space of either the haze image or the haze-free image can be disentangled into content feature space $C$ and style feature space $S$.
\end{proposition}

The difficulty of this problem lies in how to realize the complete
disentanglement of the content feature and the style feature. The
partially shared latent space assumption is common in disentangled
representation learning tasks, e.g., \cite{lee2018diverse}. It assumes that images
from different domains share one content space while keeping the style space
domain-specific. These tasks often focus on style, such as sketches or photographs,
which are rendered directly on the spatial structure of an image without editing
their content. In contrast, the style in our task is haze of different densities
leading to varying degrees of scene blocks. Thin haze will slightly lower the
visibility, while thicker haze will almost completely block the disappearance point of the
road. As mentioned before, the synthesis of haze is relatively concerned with
the depth of scenes. In other words, we need information on the scene
structure to ensure that the produced haze has the correct spatial distribution. As a result, continuing to use domain-specific style encoders will result in confusion regarding the spatial structure during the synthesis of haze, which is undesired. 

Inspired by \cite{yuxiaoming2019nips}, we align not only the content but also the style of two image domains to fully utilize all image samples. All encoders are domain-shared in our model. With the shared style encoder, the style encoder will learn to extract the style feature on the premise of understanding the semantic structure. 

Based on proposition 1, we synthesize images by mixing content features and style features extracted by encoders. Concretely, we apply the content encoder and the style encoder to extract their content features and style features, respectively:
\begin{equation}
\begin{aligned}
{c_i,s_i}=&{E^c(x_i), E^s (x_i )}, c_i\in C, s_i\in S
\\
{c_j,s_j}=&{E^c(x_j), E^s (x_j )}, c_j\in C, s_j\in S
\end{aligned}
\end{equation}

Fig.~\ref{Fig3} (a) illustrates the framework. The proposed model consists of three parts:
1) Encoders, including the content encoder $E^c$ and the style encoder $E^s$, are used to extract the content and style features of both haze images and haze-free images, respectively;
2) The only generator $G$ synthesizes images with content features and style features extracted by the encoders as input;
3) The image discriminator $D$ is used to distinguish the real images from the synthesized images and enhance the realism of the output images. Since the two image domains share the same generator for generating haze images and haze-free images, the classifier $D^{cls}$ is used to classify whether the synthesized image is a haze image or a haze-free image. The content discriminator $D^c$ takes content features as input to determine whether the content feature originates from a haze image or a haze-free image.

\subsection{Disentanglement of Content and Style}

The so-called disentanglement of content and style means that the content
feature contains no style information of the image and vice versa. The goal of complete disentanglement is to ensure that both content and style features are extracted without other redundant information. On the one hand, we use a content discriminator to eliminate style information in the content feature. On the other hand, we design a self-supervised style regression to suppress the remaining content information in the style feature.

\subsubsection{Content Feature Discriminator}
Similar to \cite{lee2018diverse}, we adopt the content feature discriminator
$D^c$ to guide the updating of the content encoder. Theoretically, it is
impossible to determine whether an image is from the haze domain or the haze-free domain based on the content feature since this feature contains no domain-related information.

First, we use the content encoder $E^c$ to extract content features $c_i, c_j$
of the input pair $x_i, x_j$. Then, we input them into the content feature
discriminator $D^c$ to determine whether the input comes from a haze image or a
haze-free image. $E^c$ is expected to extract pure content features that can
fool $D^c$, while $D^c$ tries to distinguish between the specific categories of
content features. Therefore, encoder $E^c$ and discriminator $D^c$ can
iteratively evolve through adversarial training. Once the final balance is
reached, the extracted content features do not contain
any style information of the image. The formulation of the content adversarial loss is as follows:
\begin{equation}
\begin{array}{r}
\mathcal{L}_{advc}\left(E^{c}, D^{c}\right) = \mathbb{E}_{x_{i}}\left[\log \left(D^{c}\left(E^{c}\left(x_{i}\right)\right)\right)\right] \\
+\mathbb{E}_{x_{j}}\left[\log \left(1-D^{c}\left(E^{c}\left(x_{j}\right)\right)\right)\right]
\end{array}
\end{equation}
$D^c$ tries to maximize the above loss function to classify the source of input content features. In contrast, $E^c$ tries to minimize the above loss function to fool the content classifier. Ideally, after this game process reaches the Nash equilibrium, $D^c$ cannot determine which image domain the content feature belongs to, which means that there is no haze information in the content feature. The disentanglement of style from content is successful once $E^c$ encodes only the content feature of the image.

\subsubsection{Self-supervised Style Feature Regression}



The use of the content feature discriminator and content adversarial classification loss can suppress style information in content features effectively. However, if there are no other constraints, there is no guarantee that the style feature will not contain content information.
As Fig.~\ref{Fig2} shows, the trivial content information will disturb our control process, similar to noise. Thus, to eliminate redundant content information in the style features, we propose the self-supervised style regression, as shown in Fig.~\ref{Fig3} (b). 

Representing the style encoder by a nonlinear function $f$, we first
interpolate the style features extracted from two images ($ x_i $ and $ x_j $
from the haze and haze-free domains, respectively) to obtain $s_k$:

\begin{equation}
\begin{aligned}
s_k = kf(x_i) + (1-k)f(x_j)
\end{aligned}
\end{equation}

According to the definition of disentangled representation given in \cite{higgins2018towards}, an agent's representation $f: O \rightarrow Z$ is disentangled if the map $f$ is equivariant between the actions in the observation space $O$ and those in the latent space $Z$. Such an $f$ is defined as an equivariant map or a G-morphism.


In our problem setting, the disentanglement of the style feature with respect to the content feature requires the action in the style space to correspond to that in the image space, as shown in Eq.~\ref{equ:sk}
 (the derivation can be found in the Appendix):


\begin{equation}
\begin{aligned}
s_k = f(kx_i + (1-k)x_j)
\end{aligned}
\label{equ:sk}
\end{equation}

Ideally, the interpolated $s_k$ should represent an intermediate haze density of $x_i$ and $x_j$. Thus, we combine it with the content feature $ c_j $ to synthesize an image $ x_k $. The style feature of $ x_k $ is extracted ($ s_k' $) and aligned to $s_k$ by minimizing Eq.~\ref{equ:style_reg}:

\begin{equation}
\label{equ:style_reg}
\begin{aligned}
&\mathcal{L}_s (E^c, E^s, G) = \\
&\mathbb{E}_{x_i,x_j} [\Vert E^s(G(E^c(x_j),s_k))-s_k \Vert_1
\end{aligned}
\end{equation}

Although function $f$ is nonlinear, we train it to be equivariant between the linear actions in the image space and those in the style space by random interpolation and regression. 

Assume that there exists some content information in the extracted style feature due to incomplete disentanglement in the initial stage. During training, this redundant information is interpolated, as well as style features. Since $ x_i$ and $ x_j $ have completely different content features, the interpolated content feature will be an a linear combination of two images, which is meaningless. Intuitively, the style encoder will regard the feature as
noise and eliminate it, resulting in a pure style feature $ s_k $. 

Without the monitoring of the manual label, the whole process uses $ s_k $ as a
pseudo-label to self-supervise the updating of the style encoder.
Note that this is not contradictory to our former statement about the necessity
of understanding the spatial structure for the style feature. Ideally, the
extracted style should be able to properly interpret the layout of traffic scenes so
that the synthesized haze appears in the correct place, but we do not want any traffic elements such as cars or trees to become entangled with the style feature.

\begin{figure}
	\centering
	\includegraphics[width=\linewidth]{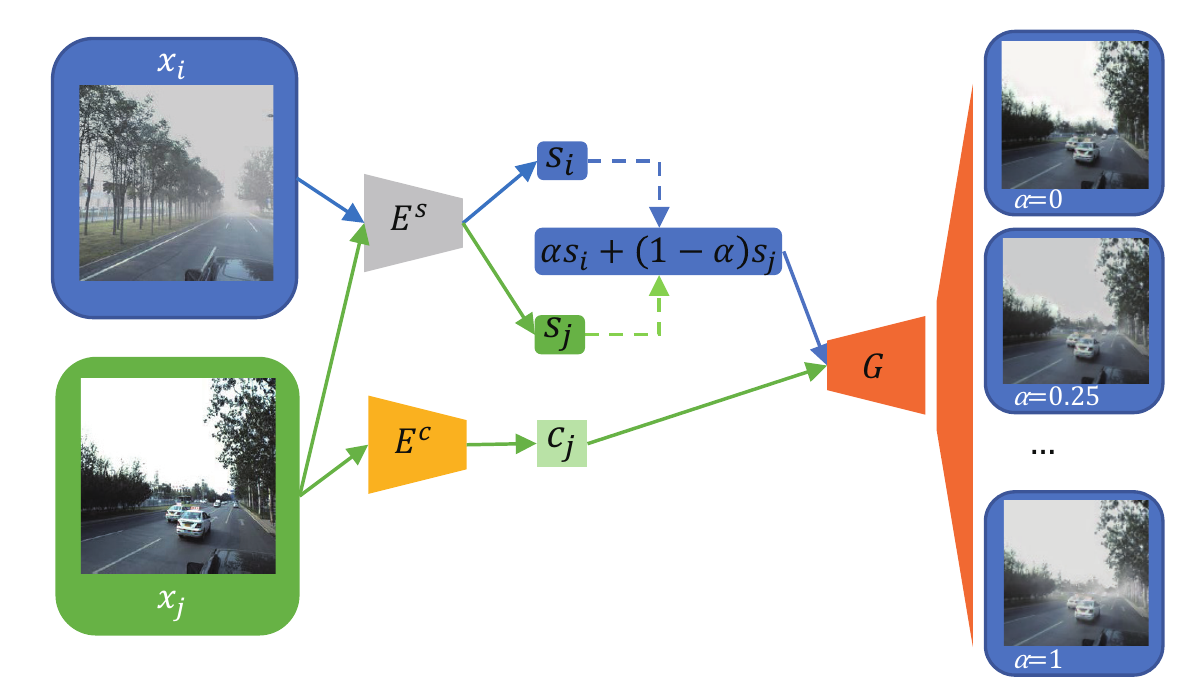}
	\setlength{\abovecaptionskip}{-5pt}
	\setlength{\belowcaptionskip}{-5pt}
	\caption{Linear style manipulation. To synthesize traffic scenes with
  controllable haze density, we extract style features from two domains
  and linearly interpolate between them with the parameter $ \alpha $. Higher $ \alpha $ is equivalent to thicker haze. Then, we combine the content feature and the interpolated style feature to produce the corresponding haze scene image with a haze density $ \alpha $.}
	\label{Fig1}
\end{figure}

Because of the random selection of $k$ during training, the network is forced to
map the haze-related style into the linear space. Therefore, we conduct linear
manipulation of the style feature and synthesize multi-density haze images as
shown in Fig.~\ref{Fig1}.


Specifically, to synthesize an image $ x_\alpha $ with a particular haze density $ \alpha $ for a certain scene, we extract the content feature $ c_j $ of the haze-free image $ x_j $ for that scene and combine it with the style feature $ s_\alpha $. $ s_\alpha $ is $ \alpha $ times $ s_i $ plus $(1-\alpha)$ times $s_j$, where $ s_i $ refers to the haze style of image $ x_i $ synthesized when the atmospheric scattering coefficient $\beta$ is equal to 1 according to Eq.~\ref{equ:haze} . The haze density is controlled by the scale factor $  \alpha$. Because $ s_i $ is derived from the baseline haze, we can scale it up or down by $ \alpha $ to obtain a scene-specific image with the desired haze concentration. The synthesis process is described by Eq.~\ref{equ:xalpha}.

\begin{equation}
\label{equ:xalpha}
\begin{aligned}
x_{\alpha} = G(E^c(x_j),{\alpha}E^s(x_i)+(1-\alpha)E^s(x_j))
\end{aligned}
\end{equation}

\subsection{Loss Function}
The overall loss function in this article contains the adversarial loss $\mathcal{L}_{adv}$, image reconstruction loss $\mathcal{L}_{recon_x}$, content reconstruction loss $\mathcal{L}_{recon_c}$, style regression loss $\mathcal{L}_{regre}$ and cycle consistency loss $\mathcal{L}_{cc}$. The complete objective function is the weighted sum of all losses:
\begin{equation}
\begin{aligned}
      \mathcal{L} =& \lambda_{adv}\mathcal{L}_{adv} + 
      \lambda_{advc} \mathcal{L}_{advc}
      + \lambda_{recon}^x \mathcal{L}_{recon}^x  
      \\
      &+ \lambda_{recon}^c \mathcal{L}_{recon}^c + \lambda_{regre}^s \mathcal{L}_{regre}^s +  \lambda_{cc}\mathcal{L}_{cc}
\end{aligned}
\label{losses}
\end{equation}
where $\mathcal{L}_{adv}=\mathcal{L}_{D_J}+\mathcal{L}_{D_I}$.
$\mathcal{L}_{D_J}$ and $\mathcal{L}_{D_I}$ are the adversarial losses in the haze
and haze-free domains, respectively. The different $\lambda$ variables are hyperparameters of the model that control the importance of each loss item.
\subsubsection{Adversarial Loss}
Adversarial loss is applied in both the haze-free image domain and the haze image domain to ensure the authenticity of the synthesized images. In the haze-free image domain, the adversarial loss is defined as follows:
\begin{equation}
\begin{aligned}
      \mathcal{L}_{D_J} = &
      \mathbb{E}_{x_j \sim P_{X_J}}\left[ logD_J(x_j) \right]\\
      + &\mathbb{E}_{x_i \sim P_{X_I}} \left[ log(1-D_J(G(E^c(x_i),s_j)))\right]
\end{aligned}
\end{equation}

$D_J$ is used to distinguish between the real haze-free image and the synthesized dehazed image; hence, $D_J$ tries to maximize the above loss function.
In contrast, $G_J$ tries to minimize the loss function to make the synthesized dehazed image more realistic.
Similarly, the adversarial loss $\mathcal{L}_{D_I}$ in the haze image domain is:
\begin{equation}          
\begin{aligned}
      \mathcal{L}_{D_I} = &
      \mathbb{E}_{x_i \sim P_{X_I}}\left[ logD_I(x_i) \right]  \\
      +&\mathbb{E}_{x_j \sim P_{X_J}} \left[ log(1-D_I(G(E^c(x_j),s_i)))\right]
\end{aligned}
\end{equation}

We value the two adversarial losses as equally influential and simply add them together to build the final adversarial loss:

\begin{equation}          
\begin{aligned}
      \mathcal{L}_{adv} = &
      \mathcal{L}_{D_J} + \mathcal{L}_{D_I}
\end{aligned}
\end{equation}

\subsubsection{Reconstruction Loss and Style Regression Loss}
Given the haze image and the haze-free image, the model should be able to reconstruct the input image after feature extraction and synthesis are completed. Therefore, the distance between the reconstructed image and the original image is used as the reconstruction loss to further constrain the model:
\begin{equation}
\begin{aligned}
      \mathcal{L}_{recon}^{x} =& 
      \mathbb{E}_{x_i \sim P_{X_I}}
      \left[\Vert G(E^c(x_i),E^s(x_i)) - x_i \Vert_1 \right] \\
      +& \mathbb{E}_{x_j \sim P_{X_J}}
      \left[\Vert G(E^c(x_j),E^s(x_j)) - x_j \Vert_1 \right] 
\end{aligned}
\end{equation}

We also hope that the content and style features of the decoded images are similar to
those of the original images. Consequently, we have the following reconstruction loss of content and style:

\begin{equation}
\begin{aligned}
      \mathcal{L}_{recon}^{c} =& 
      \mathbb{E}_{x_i \sim P_{X_I}}
      \left[\Vert E^c(G(E^c(x_i),E^s(x_i)) - E^c(x_i) \Vert_1 \right] \\
      +& \mathbb{E}_{x_j \sim P_{X_J}}
      \left[\Vert E^c(G(E^c(x_j),E^s(x_j))) - E^c(x_j) \Vert_1 \right] 
\end{aligned}
\end{equation}

\begin{equation}
\begin{aligned}
      \mathcal{L}_{recon}^{s} =& 
      \mathbb{E}_{x_i \sim P_{X_I}}
      \left[\Vert E^s_I(G(E^c(x_i),E^s(x_i))) - E^s(x_i) \Vert_1 \right] \\
      +& \mathbb{E}_{x_j \sim P_{X_J}}
      \left[\Vert E^s_J(G(E^c(x_j),E^s(x_j))) - E^s(x_j) \Vert_1 \right] 
\end{aligned}
\end{equation}

Note that we consider the style reconstruction loss to be in the same category as the
self-supervised style regression loss. We add them together with the same weights to obtain the style regression loss:
\begin{equation}
\begin{aligned}
      \mathcal{L}_{regre}^{s} = \mathcal{L}_{recon}^{s}+ \mathcal{L}_s
\end{aligned}
\end{equation}




\subsubsection{Cross-cycle Consistency Loss}
\label{cc}
We also adopt the cross-cycle consistency loss \cite{lee2018diverse} to learn the mapping between domains. For the synthesized haze image $x_{j\rightarrow i}$, the input haze-free image $x_j$ can be obtained by dehazing conversion. The cross-cycle consistency loss limits the space of the synthesized image and preserves the content characteristics of the original input image. The $l_1$ distance between the cycle reconstructed image and the input image is used as the cross-cycle consistency loss.
The forward translation process (mapping from the haze image to the haze-free image and vice versa) is as follows:
\begin{equation}
\begin{aligned}
      x_{i\rightarrow j}=G(E^c(x_i),E^s(x_j))\\
      x_{j\rightarrow i}=G(E^c(x_j),E^s(x_i))
\end{aligned}
\end{equation}
The reverse translation process (reconstructing the input image from the synthesized image) is as follows:
\begin{equation}
\begin{aligned}
      x_{i\rightarrow j \rightarrow i}=G(E^c(x_{i\rightarrow j}),E^s(x_{j \rightarrow i})) \\
      x_{j\rightarrow i \rightarrow j}=G(E^c(x_{j \rightarrow i}),E^s(x_{i \rightarrow j}))
\end{aligned}
\end{equation}
The cross-cycle consistency loss in the haze image domain and the haze-free image domain is:
\begin{equation}
\begin{aligned}
      \mathcal{L}_{cc} = \mathbb{E}_{x_i \sim P_{X_I}}\left[ \Vert x_i - x_{i\rightarrow j \rightarrow i} \Vert_1 \right]\\
                  + \mathbb{E}_{x_j \sim P_{X_J}}\left[ \Vert x_j - x_{j\rightarrow i \rightarrow j} \Vert_1 \right]
\end{aligned}
\end{equation}

\section{Experiments}

To verify the effectiveness of the method presented in this paper, this section analyses the impact of different modules or losses on the synthesized results and compares them with those of the existing methods both quantitatively and qualitatively.
This section first introduces the datasets and implementation details of the method,
including a detailed introduction to the network structure and
hyperparameters. Then, we provide a detailed analysis of our model, including
a comparison with the existing methods. In addition, visualizations of intermediate results and generalization analysis are presented to further clarify our method’s effectiveness. The last part of this section details the ablation
studies. All experiments are conducted on an NVIDIA TITAN XP GPU with 12 GB of memory.

\subsection{Dataset}
The dataset is synthetic. Specifically, we use haze-free images (4000 randomly selected images) and corresponding depth images from the ApolloScape dataset \cite{Huang2018The} to synthesize haze images according to the atmospheric scattering model. For a given haze-free image and its depth image, we synthesize the corresponding haze image according to equation \ref{equ:haze}. We randomly set the atmospheric light intensity $A$ from $0.8$ to $1.0$ and the atmospheric scattering coefficient $\beta$ to 1. Finally, we select 1500 clear images and 1500 synthetic haze images as the training data and 500 of each as the test data.

\subsection{Implementation Details}
The network of our proposed model consists of a content encoder, a style
encoder, a generator, a discriminator and a content adversarial classifier.
The encoders have the same structure as that in \cite{lee2018diverse}. The content encoder
consists of three convolutional layers and four residual blocks. The style
encoder consists of four convolutional layers and a fully connected layer. The
structure of the generator and the content encoder is symmetrical, consisting of
four residual blocks and three transposed convolutions. Both the discriminator and
the content adversarial classifier adopt the
PatchGAN structure \cite{Isola2016Image}, which is composed of four convolutional layers
and a discriminator with a multiscale structure. During training, we use
mini-batch stochastic gradient descent (batch size is 4) and the Adam optimization
algorithm ($\beta_1=0.5$, $\beta_2=0.999$). The initial learning rate is 0.0001,
and the learning rate is reduced by half every 10,000 iterations. In all the experiments, the images are randomly cropped to a size of $256 \times 256$ as the input. The model hyperparameters are specifically set as follows: $\lambda_{adv} =1, \lambda_{advc}=10, \lambda_{recon}^x=10, \lambda_{recon}^c=1, \lambda_{regre}^s=20,$ $\lambda_{cc}=10$.

The detailed network structure is shown in Table~\ref{archi}.

\begin{table}[htpb]  
	
	\caption{Network structure. N, K, and S respectively represent the number of channels of the convolutional layer, the size of the convolution kernel, and the stride of the convolution. RESBLK stands for the residual block (Residual Block).}
	\label{archi}
	\begin{center}
	\begin{tabular}{p{25pt}p{100pt}p{65pt}}  
		\toprule   
		Layer & Content Encoder & Layer Info. \\  
		\midrule   
		1 & CONV & N64, K7, S1    \\  
		
		2 & CONV, RELU & N128, K4, S2   \\ 
		
		3 & CONV, RELU & N128, K4, S2   \\ 
		
		4 & RESBLK, RELU &N256, K3, S1 \\
		
		5 & RESBLK, RELU &N256, K3, S1 \\
		
		6 & RESBLK, RELU &N256, K3, S1 \\
		
		7 & RESBLK, RELU &N256, K3, S1 \\
		\bottomrule  
	\end{tabular}
	
	\vspace{3pt}
	
	\begin{tabular}{p{25pt}p{100pt}p{65pt}}  
		\toprule   
		Layer & Style Encoder &Layer Info. \\  
		\midrule   
		1 & CONV, RELU & N64, K7, S1    \\  
		
		2 & CONV, RELU & N128, K4, S2  \\    
		
		3 & CONV, RELU & N256, K4, S2  \\    
		
		4 & AdaptiveAvgPool & Output=1\\
		
		5 & CONV, Sigmoid & N8, K1, S1  \\
		\bottomrule  
	\end{tabular}
	
	\vspace{3pt}
	
	\begin{tabular}{p{25pt}p{100pt}p{65pt}}  
		\toprule   
		Layer & Generator &Layer Info. \\  
		\midrule   
		1 & RESBLK, RELU &N256, K3, S1 \\
		
		2 & RESBLK, RELU &N256, K3, S1 \\
		
		3 & RESBLK, RELU &N256, K3, S1 \\
		
		4 & RESBLK, RELU &N256, K3, S1 \\
		
		5 & DECONV, RELU &N128, K5, S1 \\
		
		6 & DECONV, RELU &N64, K5, S1 \\
		
		7 & DECONV, RELU &N3, K7, S1 \\
		\bottomrule  
	\end{tabular}
	\end{center}

\end{table}

\subsection{Performance Evaluation}
In this section, we provide a qualitative comparison of our model results with
the state-of-the-art (SOTA) results and an in-depth analysis of the disentanglement
effects, concluding with a description of the role and irreplaceability of each
submodule based on ablation studies.

\begin{figure*}
	\centering
	\includegraphics[width=\linewidth]{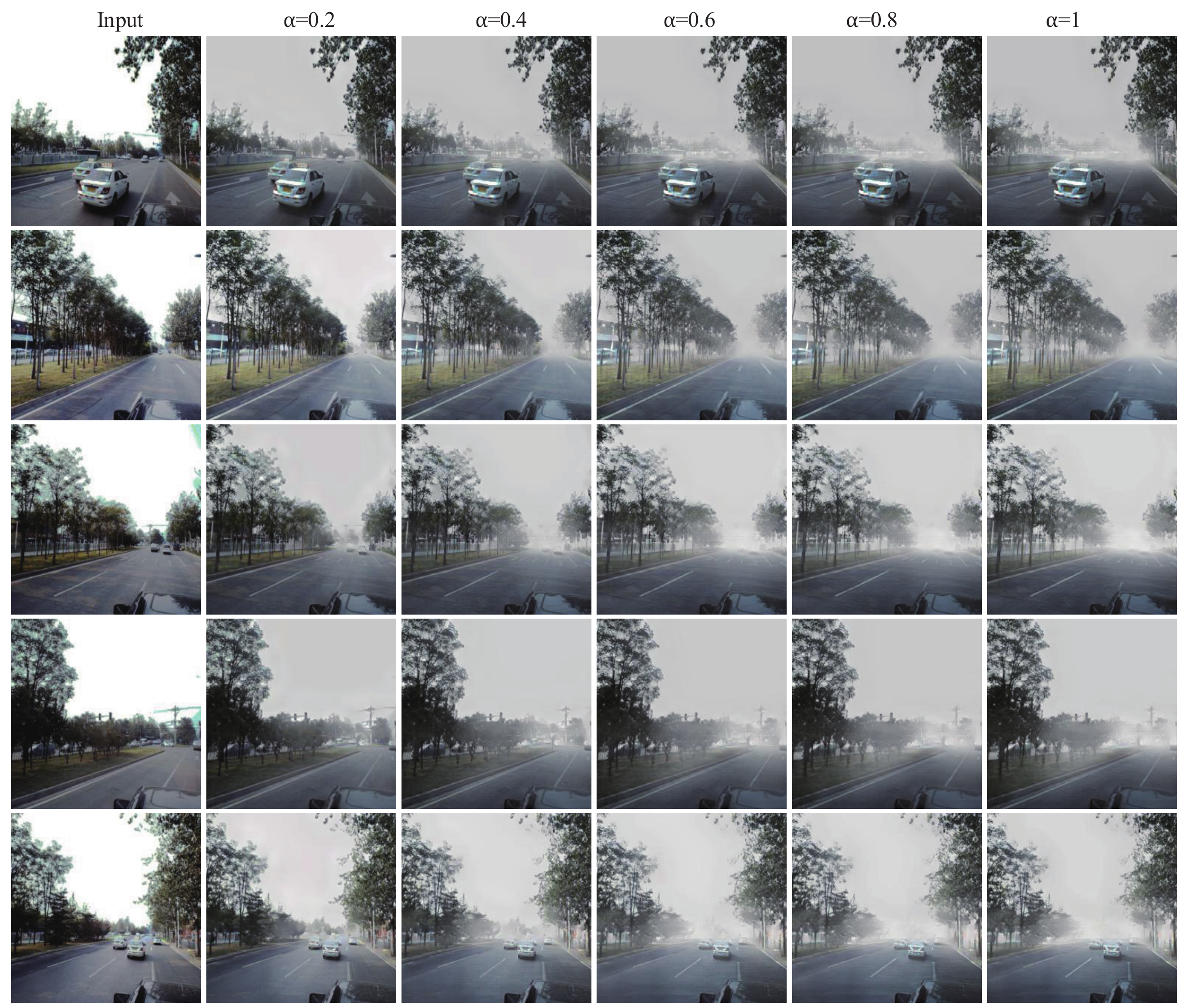}
	\setlength{\abovecaptionskip}{-5pt}
	\setlength{\belowcaptionskip}{-5pt}
	\caption{Multi-density results synthesized by controlling the parameter $\alpha$. From the left to right column, vehicles and trees near the vanishing points of the road gradually become invisible as $\alpha$ increases.}
	\label{Fig6}
\end{figure*}

\subsubsection{Evaluation Metrics}
\label{evaluation}

We first evaluate the image synthesis quality from the aspect of traditional
computer graphics: peak signal-to-noise ratio (PSNR) and structural similarity
index (SSIM). Evaluation metrics based on the image depth features are also used, e.g., the Fréchet inception distance (FID) \cite{heusel2017gans}, LPIPS distance \cite{Zhang2018The} and
VGG distance.

The PSNR is based on the error between corresponding pixel points and is one of the most widely used objective evaluation metrics for images. Larger values indicate less distortion of the synthesized images.

The basic idea of the SSIM is to evaluate the similarity between two images from three aspects: brightness, contrast and structure. The SSIM takes values in the range [0, 1], with larger values indicating less image distortion.

The FID is the calculated Fréchet distance of two sets of images
on the feature of the inception network \cite{szegedy2016rethinking}, which is used
to evaluate the similarity of the two sets of images. We use it to measure the
similarity between the synthesized images and reference images. Lower FID values
mean that the synthesized images are closer to the reference images at the image level.

The LPIPS distance, which is obtained by calculating the weighted Euclidean distance between the depth features of a pair of images, is often utilized to measure diversity. Higher LPIPS scores indicate better diversity among the synthesized images. Similar to \cite{zhu2017toward}, the experiments in this paper calculate the average LPIPS distance between randomly sampled image pairs from synthesized haze images to evaluate their diversity.

The VGG distance is the distance between the VGG \cite{Simonyan2014Very} feature of the input image and that of the synthesized image. Different from the FID, the VGG distance can better evaluate the similarity between the synthesized image and the ground truth at the semantic level. When calculating the VGG distance, we use the output of the $pool5$ layer of the VGG-19 network as the image feature.

\subsubsection{Results}

\begin{figure*}
	\centering
	\includegraphics[width=\linewidth]{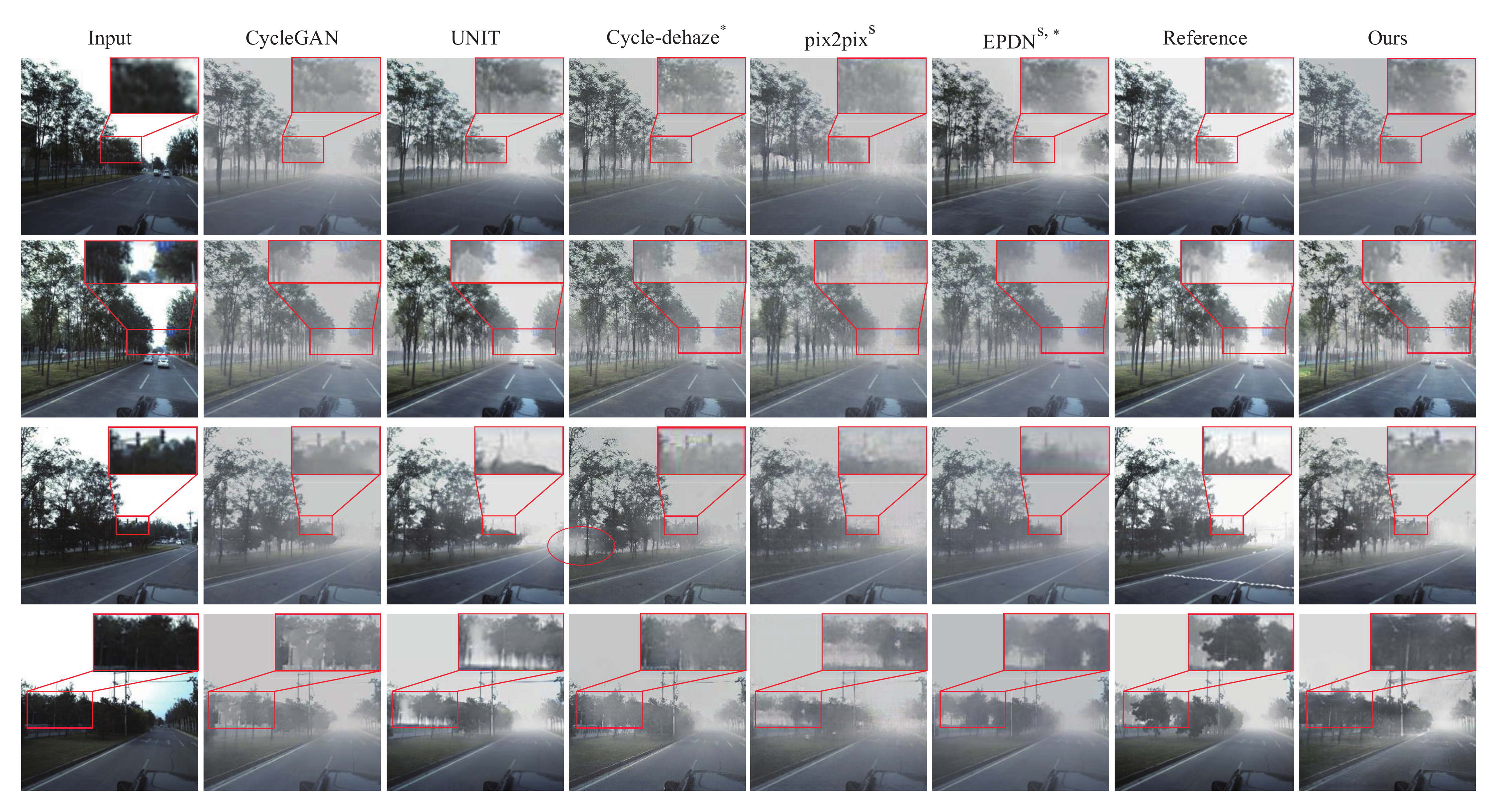}
	\setlength{\abovecaptionskip}{-5pt}
	\setlength{\belowcaptionskip}{-5pt}
	\caption{Synthesized haze images of our method, compared with mainstream image-translation methods and relevant SOTA dehazing methods. Columns marked with an ``s'' are supervised methods, while the rest are unsupervised methods. In comparison, those with an ``$*$'' (the fourth and sixth columns) are SOTA dehazing models, which can synthesize haze in reverse; and the rest (the second, third and fifth columns) are mainstream image-translation models. The ``Reference" column refers to images synthesized via the atmospheric scattering model (Eq.~\ref{equ:haze}). }
	\label{Fig7}
\end{figure*}

In this subsection, we synthesize multi-density haze images by controlling the aforementioned parameter $\alpha$ and compare the proposed model with the relevant SOTA methods. 

Images with diverse haze densities are shown in Fig.~\ref{Fig6}. Note that we
consider the value corresponding to haze that has just emerged to be 0 and the
baseline haze to be 1, which refers to haze synthesized with the atmospheric
scattering coefficient $ \beta $ equal to 1. Then, we scale the
haze feature up and down in the interval from 0 to 1. From the results, we can tell the
difference in haze density among images with different values of the
parameter $\alpha$, which indicates that we successfully achieve the
ability to distinguish produced haze by controlling $ \alpha $. For example,
vehicles and trees near vanishing points of the road gradually become
indistinguishable as $\alpha$ increases. Under the influence of the style
feature regression, our networks memorize the distinction between thick and thin haze and learn to synthesize haze images without sacrificing realism.

The qualitative comparison result is shown in Fig.~\ref{Fig7}. In this experiment, we compare our synthesized haze images with mainstream image-translation methods (CycleGAN \cite{Zhu2017Unpaired}, UNIT \cite{Liu2017Unsupervised}, pix2pix \cite{Isola2016Image}) and SOTA dehazing methods (Cycle-dehaze \cite{engin2018cycle}, enhanced pix2pix dehazing network (EPDN) \cite{qu2019enhanced}). It is worth to note that these two methods can also synthesize haze since they are formulated as an image-translation process. The training images of all models come from the same dataset but are organized differently. Supervised models like EPDN \cite{qu2019enhanced} and pix2pix \cite{Isola2016Image} are trained on paired images, while the rest unsupervised models are trained on unordered data.

From the results, we can see that other unsupervised models like CycleGAN \cite{Zhu2017Unpaired}, UNIT \cite{Liu2017Unsupervised} and Cycle-dehaze \cite{engin2018cycle} mainly suffer from two defects. In the first and second rows of Fig.~\ref{Fig7}, clear discontinuity appears at the boundary of the haze. Besides, the jagged haze boundaries reduce the realism of the synthesized images. %
In contrast, the results of this paper own gradually disappearing haze boundaries, which indicate a higher image quality. It is because that our method abstracts the haze style from the content to interpret the characteristics of the haze. Therefore, the continuous and realistic haze is synthesized, ignoring existing errors. In the third and fourth rows, the visual appearance of the image content is corrupted to varying degrees. For example, other methods produce truncated trees (as shown in the third row) or trees blocked by irregular haze (as shown in the fourth row and the red circular area), whereas our method preserves the structure of the trees in the input image.

In most cases, supervised models can achieve better results than unsupervised models since they obtain more pixel level correspondence between two domains during the training stage. However, as an unsupervised approach, the results of our method are comparable to those of supervised frameworks such as pix2pix \cite{Isola2016Image} and the EPDN \cite{qu2019enhanced} and even better in terms of the details. For example, there exists unnatural grid-like stripes in the synthesized images of pix2pix \cite{Isola2016Image}. The EPDN \cite{qu2019enhanced} applies an extra enhancer on the basis of pix2pixHD \cite{wang2018high} and eliminates such stripes, but there still occurs an unbalanced distribution of synthesized haze (as shown in the zoomed-in areas). Note that, benefiting from paired training data, the supervised models could synthesize haze images that are the closest to the reference images. Nevertheless, a closer distance to the reference image does not necessarily mean a better synthetic quality in our case because the reference images are synthesized based on the atmospheric scattering model with imperfect depth information. Since both pix2pix \cite{Isola2016Image} and the EPDN \cite{qu2019enhanced} learn the mapping relationship between image domains at the pixel level, their results are inevitably affected by errors existing in the depth images. In contrast, our approach learns the haze distribution from the reference images as well as the content distribution from the haze-free images. Under more relaxed training constraints, our approach can better incorporate content features to synthesize evenly distributed haze.



%

\begin{table}[!tbh]
	\setlength{\abovecaptionskip}{0pt}
	\setlength{\belowcaptionskip}{0pt}
	\caption{Quantitative comparison with mainstream image-translation methods and relevant SOTA dehazing methods. $d_{\text{VGG}}$ and $d_{\text{LPIPS}}$ denote the \text{VGG} distance and \text{LPIPS} distance, respectively. Among them, the smaller the value of $d_{\text{VGG}}$ and the FID, the better the experimental result.
The larger the values of the other indicators, the better the experimental result. The boldface figures indicate the best values. The supervised methods are marked with an ``s'', while the rest are unsupervised models. Those names with an asterisk are SOTA dehazing models, which can synthesize haze in reverse, and the rest are mainstream image-translation models.}
	\label{tab:base}
	\centering
	\begin{tabular}{cccccc}
		\toprule
		\textbf{module} & \textbf{SSIM}$\uparrow$ & \textbf{PSNR}$\uparrow$ & \textbf{$d_{\text{VGG}}$}$\downarrow$ & \textbf{FID}$\downarrow$  & \textbf{$d_{\text{LPIPS}}$}$\uparrow$\\
		\midrule
		\textbf{UNIT} & 0.552 & 17.518 & 6.452 & 28.138 & 0.064 \\
		\textbf{CycleGAN} & 0.586 & 17.192 & 6.726 & 29.665 & 0.052\\
		\textbf{Cycle-dehaze$^*$} & 0.590 & 19.907 & 4.550 & \textbf{26.070} & 0.068\\
		\textbf{pix2pix$^s$} & 0.698 & 18.736 & 4.418 & 27.084 & 0.043\\
		\textbf{EPDN$^{s,*}$} & \textbf{0.854} & \textbf{24.949} & \textbf{3.777} & 30.874 & 0.066\\
		\textbf{Ours} & 0.769 & 18.051 & 4.058 & 26.311 & \textbf{0.072}\\
		\bottomrule
	\end{tabular}
\end{table}

Using the evaluation metrics described in Sec.~\ref{evaluation}, we conduct
comparison experiments with the classical approaches, as shown in
Table~\ref{tab:base}. From the perspective of image quality (SSIM, PSNR,
VGG distance, FID), our method performs best in terms of almost all metrics compared with the mainstream image-translation methods (UNIT \cite{Liu2017Unsupervised},
CycleGAN \cite{Zhu2017Unpaired}, pix2pix \cite{Isola2016Image}). This indicates that our approach produces higher-quality images compared with those of the mainstream image-translation models. 

We further compare our method with relevant SOTA dehazing models. The proposed approach outperforms in terms of the SSIM and VGG distance and has the same level of performance considering the FID compared with the unsupervised Cycle-dehaze method \cite{engin2018cycle}. Note that,  the SOTA dehazing model EPDN is superior in terms of the SSIM, PSNR and VGG distance, as expected. However, the EPDN learns the mapping relationship between image domains at the pixel level, its results are inevitably affected by the error existing in depth images (as illustrated in Fig.~\ref{Fig7}). In contrast, our approach learns the haze distribution from the reference images as well as the content distribution from the haze-free images. Under more relaxed training constraints, our approach can better incorporate content features to synthesize equally distributed haze. Moreover, our results are even better considering the FID, which means that the haze images synthesized by our model are more realistic.

In terms of diversity, our method is superior to all other methods on the basis of the LPIPS distance. This proves that our algorithm can synthesize images that are not only realistic but also the most diverse.

\subsubsection{Disentanglement Analysis}
\begin{figure*}
	\centering
	\includegraphics[width=\linewidth]{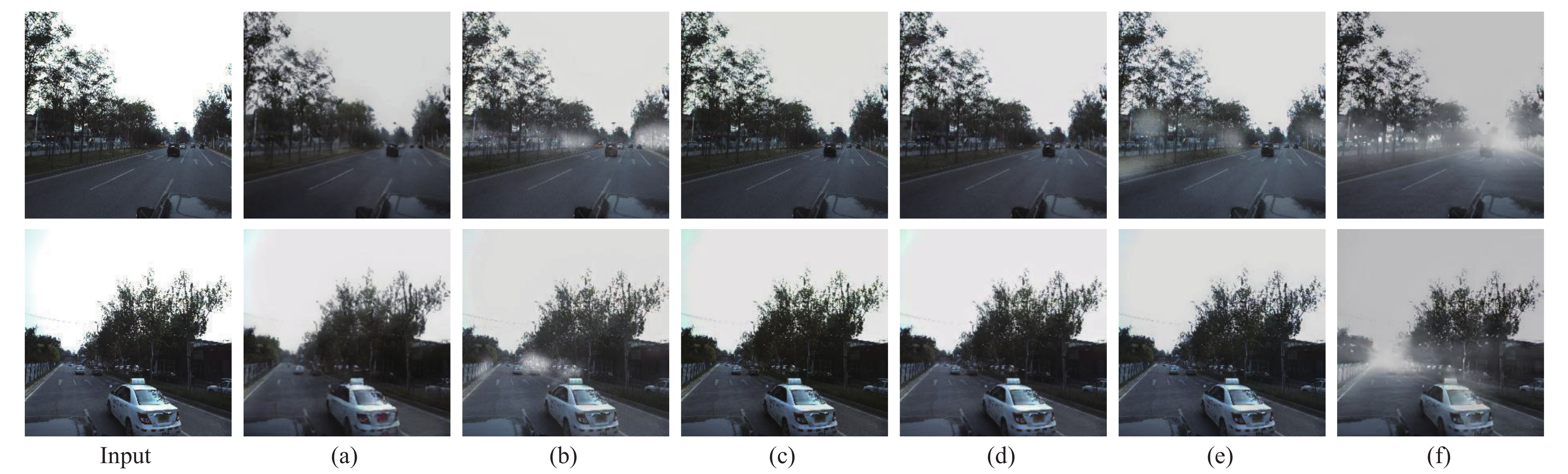}
	\caption{Qualitative comparison when different modules are removed. Columns (a) to (f) represent the result of removing $\mathcal{L}_{recon}^x$, $\mathcal{L}_{recon}^c$, $\mathcal{L}_{advc}$, $\mathcal{L}_{cc}$, $\mathcal{L}_{regre}^s$ and the result of using the complete model, respectively. The incomplete models either fail to synthesize haze as (f) does or produce only separated patches of haze.}
	\label{Fig5}
\end{figure*}
Although the content features and style features of images are extracted by the content encoder and style encoder, respectively, and we apply a few constraints to suppress the redundant information in each, it is not explicit enough to judge whether the model successfully realizes disentanglement and how completely it does so. Separately observing the extracted feature statistics is useless when attempting to
explain how well the disentanglement works. Instead, we consider what happens if the disentanglement is not effective. Ideally, the content feature of an image should not contain any style information that represents a certain
density of haze. In other words, it is impossible to classify whether an image is a haze-free image or a haze image based only on its content feature if the content feature does not entangle with any style knowledge. On the other hand,
the style feature can be used to directly classify whether the real image belongs to the haze image or the haze-free image. Therefore, we train a classification network with different inputs.   

Specifically, we evaluate the classification accuracy on both the train set and the test set. After the training process, the learned features are sent into the classification network to judge whether the input feature comes from the haze domain or the haze-free domain. The experimental results are shown in Table~\ref{tab:dis}. Ideally, it is difficult for the classifier to classify whether an image is a haze-free image or a haze image only based on its content feature if the content feature is disentangled. In other words, the closer the classification accuracy of the content feature is to 50$\%$, the more complete the disentanglement is.

Trained and tested on such small data, a neural network classifier should be able to achieve a high accuracy on the binary classification problem if the input features are distinctive. However, the classifier only achieves an accuracy of 61.5$\%$ on the train set when applying content features, which already shows that the classifier failed to find the distinct information. Furthermore, the accuracy on the test set is closer to 50$\%$, proving the classifier's inability to discriminate from the content features. The classifier performs consistently on both the training and test sets, confirming the success of our model in disentangling style and content. The classifier performs consistently on both the training and test sets,  confirming the success of our model in disentangling style and content.

\begin{table}[!tbh]
	\setlength{\abovecaptionskip}{0pt}
	\setlength{\belowcaptionskip}{0pt}
	\caption{Classification accuracy in different feature spaces.}
	\label{tab:dis}
	\centering
	\begin{tabular}{ccccc}
		\toprule
		\textbf{Accuracy} & \textbf{Content} & \textbf{Style} & \textbf{CycleGAN} & \textbf{Image}\\
		\midrule
		\textbf{training set}& 0.615 & 0.987 & 0.875 & 0.973 \\
		\textbf{test set} & 0.581 & 0.972 & 0.853 & 0.952 \\
		\bottomrule
	\end{tabular}
\end{table}

\subsection{Further Analysis}
To illustrate the roles of the various modules of our model and their irreplaceability, we draw on several ablation experiments in the following.
\subsubsection{Quantitative Evaluation}

\begin{table}[!tbh]
	\setlength{\abovecaptionskip}{0pt}
	\setlength{\belowcaptionskip}{0pt}
	\caption{Quantitative model comparison results obtained after removing different losses. We demonstrate the effect of the content discriminator, cross-cycle consistency loss, reconstruction losses and style regression loss. Among them, the smaller the value of $d_{\text{VGG}}$ and the FID, the better the experimental result is.
The larger the values of the other indicators, the better the experimental result is. The boldface figures indicate that the value is the best.}
	\label{tab:compare}
	\centering
	\begin{tabular}{cccccc}
		\toprule
		\textbf{Module} & \textbf{SSIM}$\uparrow$ & \textbf{PSNR}$\uparrow$ & \textbf{$d_{\text{VGG}}$}$\downarrow$ & \textbf{FID}$\downarrow$  & \textbf{$d_{\text{LPIPS}}$}$\uparrow$ \\
		\midrule
		\textbf{removing \text{$\mathcal{L}_{cc}$}}  & 0.670 & 12.922 & 5.950 & 68.863 & 0.011 \\
		\textbf{removing \text{$\mathcal{L}_{advc}$}} & 0.645 & 12.238 & 6.300 & 74.659 & 0.012 \\
		\textbf{removing \text{$\mathcal{L}_{recon}^x$}} & 0.588 & 13.267 & 4.688 &  50.246 &  0.023 \\
		\textbf{removing \text{$\mathcal{L}_{regre}^s$}} & 0.676 & 13.256 & 6.080 & 70.969 & 0.012 \\
		\textbf{removing \text{$\mathcal{L}_{recon}^c$}}  & 0.697 & 14.388 & 5.496 & 50.234 & 0.029 \\
		\textbf{complete model} & \textbf{0.769} & \textbf{18.051} & \textbf{4.058} & \textbf{26.311} & \textbf{0.072}\\
		\bottomrule
	\end{tabular}
\end{table}

To further understand the impact of the various modules of the method in this paper, 
we evaluate the synthesis results of the incomplete models with certain modules removed and compare them with those of the complete model.
Table~\ref{tab:compare} shows the quantitative results obtained using the metrics
mentioned in Sec.~\ref{evaluation}. The complete model performs far better on all metrics. Fig.~\ref{Fig5} shows the haze images synthesized by
these models. From the figure, the incomplete
models all have difficulty in generating realistic haze, which further validates the effectiveness of each separate module.

\subsubsection{Ablation Study on the Self-supervised Style Regression}
\begin{figure*}
	\centering
	\includegraphics[width=\linewidth]{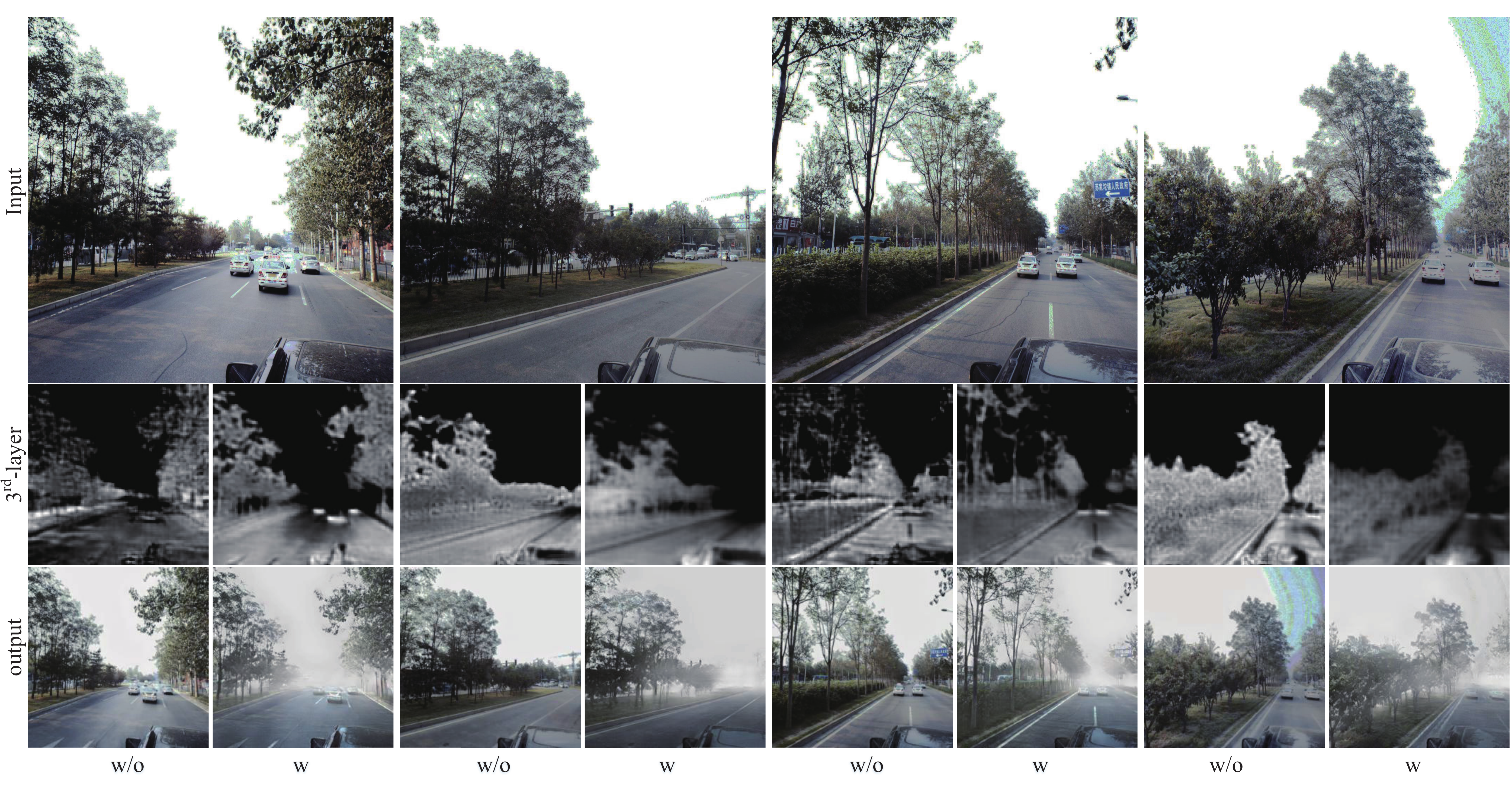}
	\setlength{\abovecaptionskip}{-5pt}
	\setlength{\belowcaptionskip}{-5pt}
	\caption{Visual comparisons of the latent transmittance map with and without the process of self-supervised style regression. The first row presents the input images. The images in the second row are the outputs of the third layer of our generator, and the corresponding results are given in the last row. Vertically, columns with the index ``w/o" are experimental results obtained without self-supervised style regression, while the ``w" columns are the opposite.}
	\label{Fig-wointerloss}
\end{figure*}
As previously mentioned, the transmittance map $ t(x) $ of the image plays a
major role in the haze rendering process. For our extracted style features to be
a certain form of the mapping and to not contain redundant content information,
we propose self-supervised style regression. Fig.~\ref{Fig-wointerloss}
visualizes the impact of this operation on the feature maps learned by the
network. The pictures of the $ 3^{rd} $-layer are the outputs of the third
layer of the generator. We can see from the pictures that, after inputting the
extracted style features into the generator, they can be decoded into such
feature maps, which are similar to the transmittance map of the scenes. This
indicates that the style feature we extracted can be viewed as the latent
transmittance map of the input images. It is obvious from the results that in the
absence of self-supervised style regression, the feature maps obtained from the
decoding of the style feature give a clear description of the entire
scene content, which indicates that the style feature extracted by the network
also contains redundant content information, such as the edges of the scene. The quality of the synthesized haze is hence compromised. Conversely, the scene structure in the feature map becomes blurred under the constraints of self-supervised regression. This shows that the style encoder focuses more on extracting the haze-related style feature and reducing the extracted content information when encoding and further proves the effectiveness of our method.

\subsubsection{Ablation Study on the Shared Style Encoder}

\begin{figure}[!t]
	\centering
	\setlength{\abovecaptionskip}{0pt} 
	\setlength{\belowcaptionskip}{0pt} 
	\includegraphics[width=0.5\textwidth]{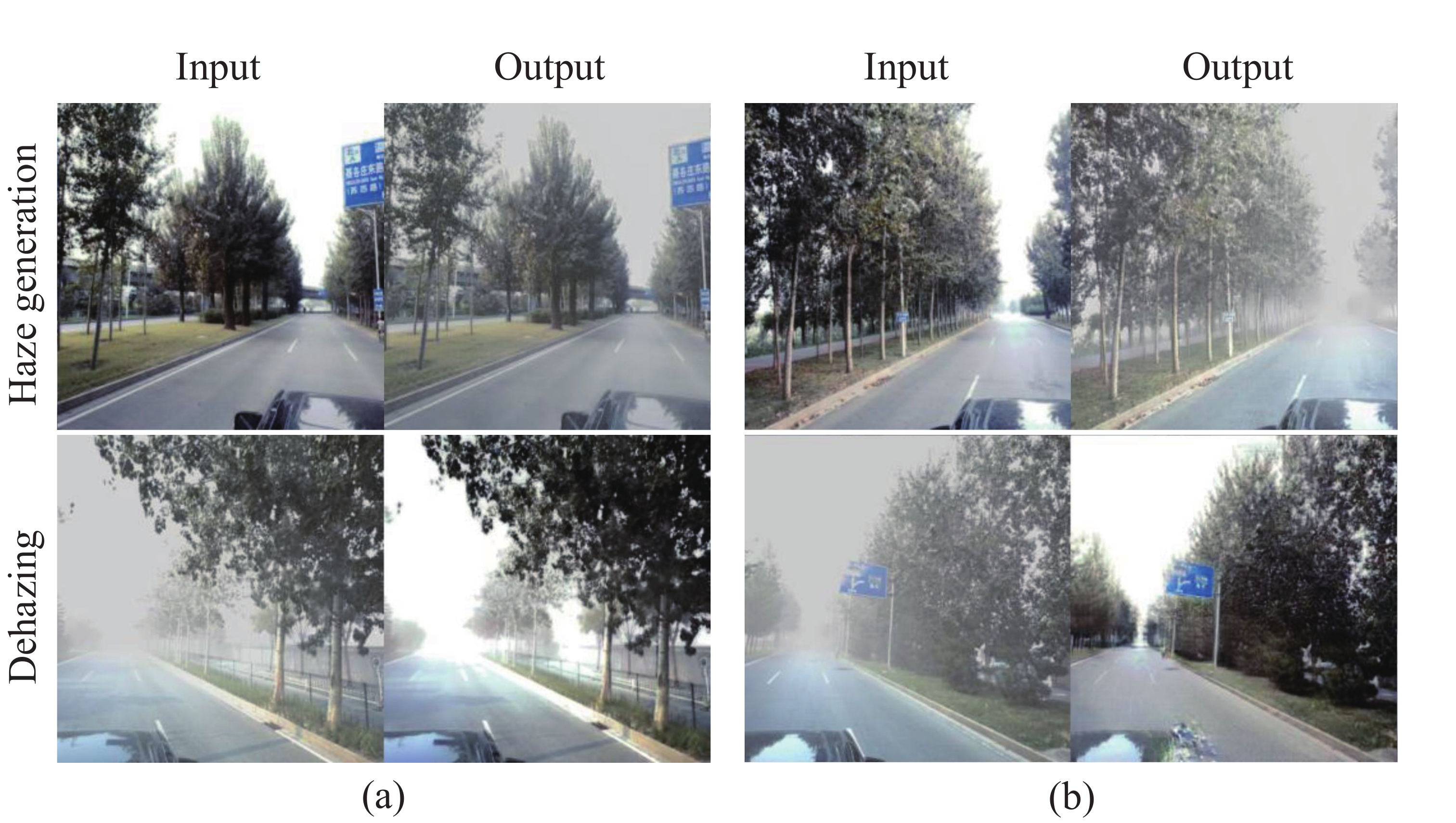}
	\caption{Visual comparisons between (a) the model with domain-specific style encoders and (b) the model with a shared style encoder. The first row shows images of the process of adding haze, while the second row shows the outputs of dehazing.}
	\label{fig_2encoders}
\end{figure}
To ensure that the resulting haze is distributed in the right places, we use a shared style encoder in our model. We hope that the haze-free images can provide an essential understanding of the spatial structure of scenes to the haze-related style features extracted from haze images. To verify this, we set up the following experiment for comparison: two separate style encoders are used to extract the style feature from the haze-free images and haze images. 

The results are shown in Fig.~\ref{fig_2encoders}. From the images, we can tell that compared to using a shared style encoder, the model with domain-specific style encoders is able to add only a uniform luminance, which indicates that the disentanglement of content and style is degraded. Due to this incomplete disentanglement, parts of the prior knowledge of the scenes that the content encoder should have learned are missing; thus, the model fails to synthesize elements at the vanishing point of the road in the dehazing process. In contrast, applying a shared style encoder to both image domains ensures that the encoders fully utilize all samples from the two domains and achieve successful disentanglement.

\subsubsection{Visualization of the Content Features}

\begin{figure}[!t]
	\centering
	\setlength{\abovecaptionskip}{5pt} 
	\setlength{\belowcaptionskip}{0pt} 
	\includegraphics[width=0.5\textwidth]{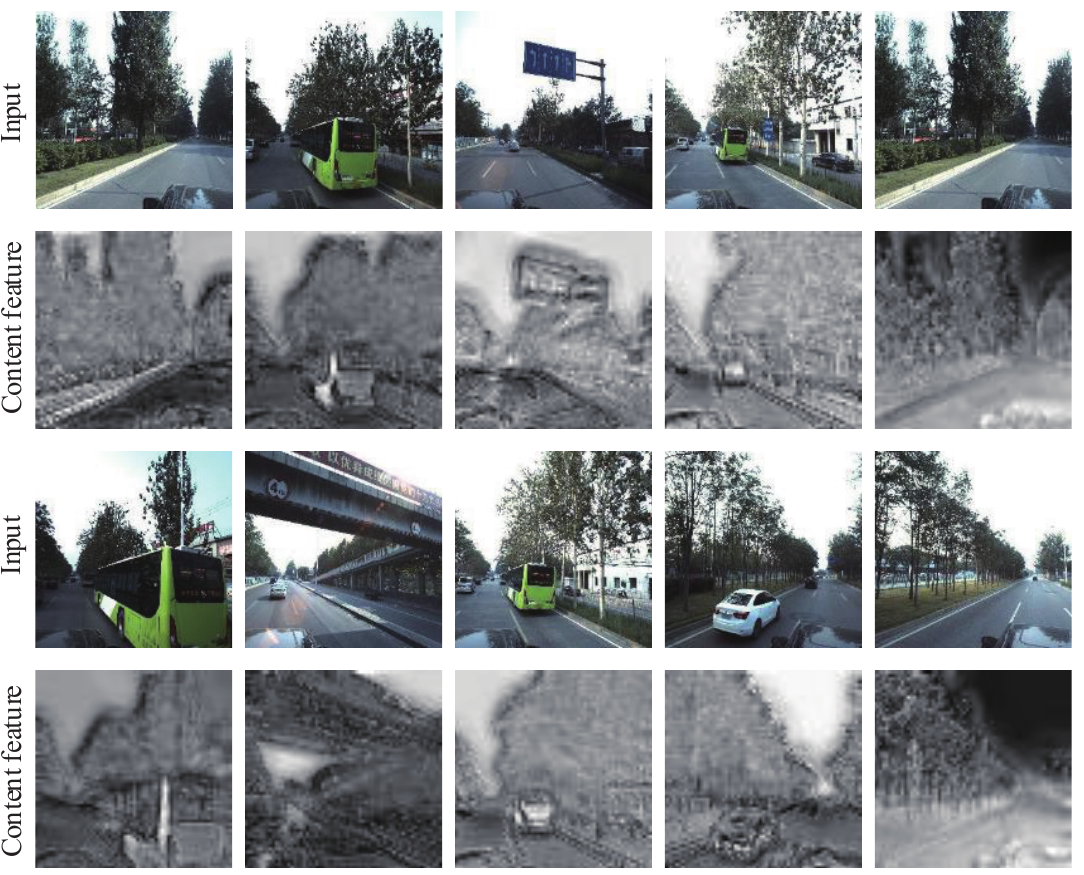}
	\caption{{\color{black}{Input images and their visualized content features. The low-level structures of traffic elements such as vehicles, traffic signs and trees are well preserved in the disentangled content features.}}}
	\label{content}
\end{figure}

We visualized all the content features of the test sets as shown in Fig.~\ref{content}. For a single image, its content feature is a tensor whose shape is [256, 64, 64], from which we selected the first value to perform the visualization without loss of generality. As we can see, the low-level structures of traffic elements are well preserved in the disentangled content features, including vehicles, traffic signs, flyovers and trees. Information related to tiny details such as edges and lanes are also extracted well. This empirical evidence verifies that the content information describing the semanteme of different objects is well extracted by the content encoder. 

\subsubsection{Visualization of the Intermediate Results}
\begin{figure*}
	\centering
	\includegraphics[width=\linewidth]{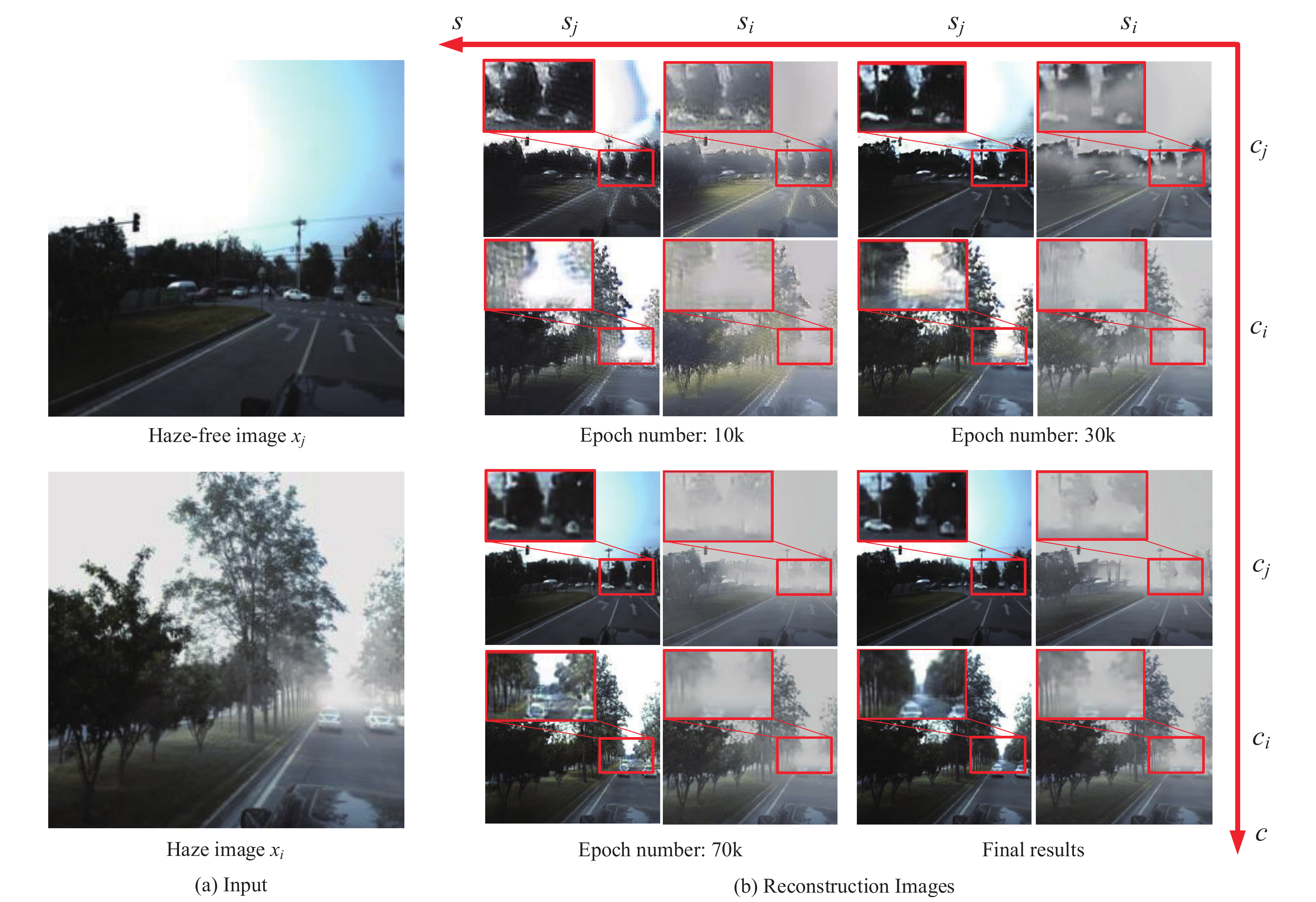}
	\setlength{\abovecaptionskip}{-5pt}
	\setlength{\belowcaptionskip}{-5pt}
	\caption{The visualization result of images reconstructed in different training epochs. (a) The input haze image $x_i$ and the haze-free image $x_j$. (b){\color{black}{ The reconstructions from the style-content feature pairs, i.e., $s_i$-$c_i, s_i$-$c_j, s_j$-$c_i$ and $s_j$-$c_j$ , at different training epoch numbers. $s_i$ and $c_i$ are style features and content features extracted from the input haze images $x_i$ while $s_j$ and $c_j$ are that of haze-free images $x_j$.}}}
	\label{recons}
\end{figure*}

Fig.~\ref{recons} shows the reconstructions from the style-content feature pairs, i.e., $s_i$-$c_i, s_i$-$c_j, s_j$-$c_i$ and $s_j$-$c_j$, at different training times (i.e., different epoch numbers).

As we can see from the zoomed-in parts, the quality of reconstructed images improves significantly with more iterations. Specifically, at 10k and 30k, the synthesized haze in the reconstructed haze images is unnatural and unrealistic. The semantic content in the reconstructed haze-free images also suffers from the lack of details near the vanishing point of the road. However, as the training process continues, our model gradually learns to synthesize realistic haze and reconstruct the missing content details, which proves that our learning strategy is effective.

\subsubsection{Visualization of the Changes in Loss}

\begin{figure*}
	\centering
	\includegraphics[width=\linewidth]{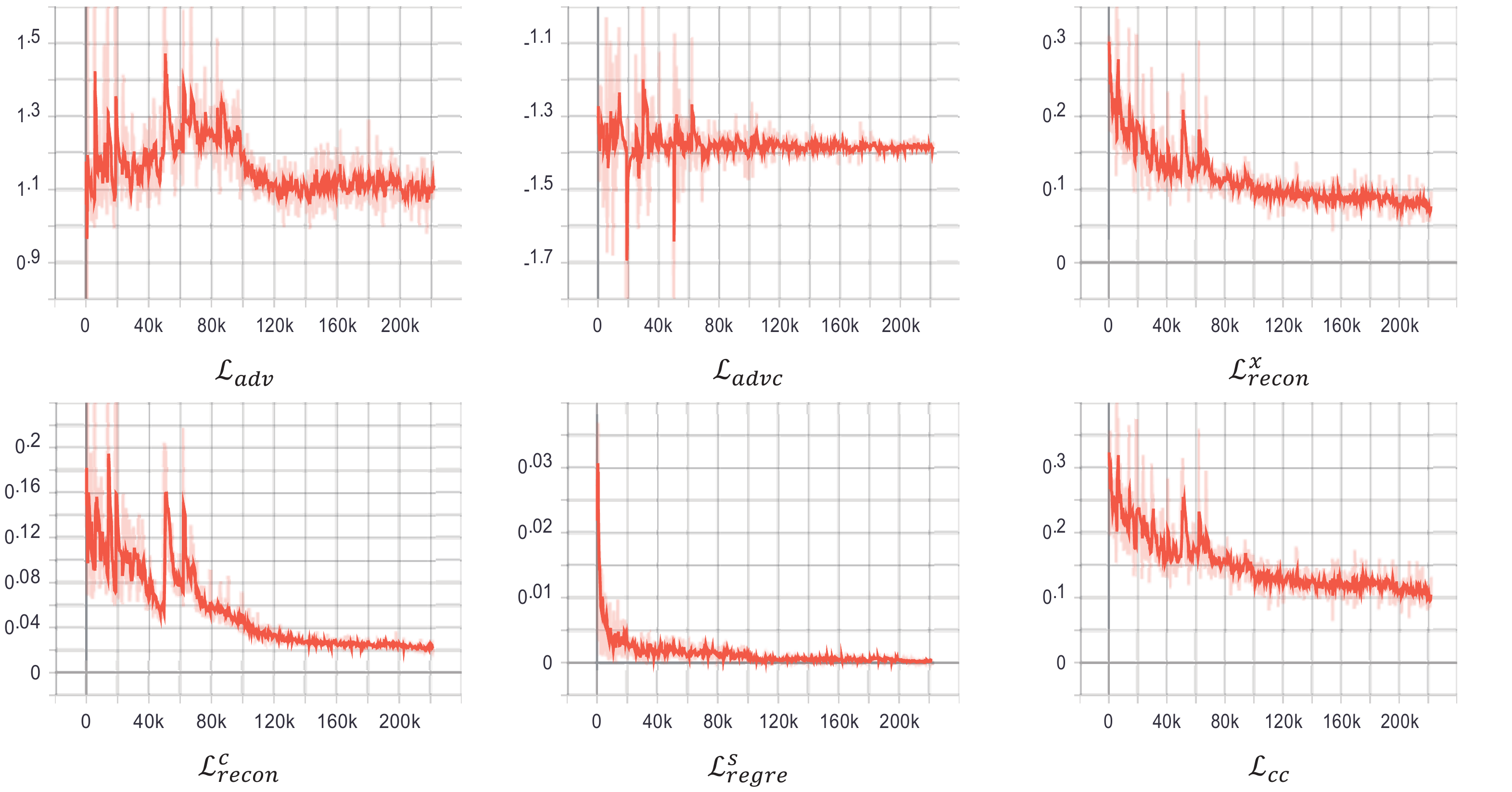}
	\setlength{\abovecaptionskip}{-5pt}
	\setlength{\belowcaptionskip}{-5pt}
	\caption{{\color{black}{Learning curves of each loss described in Eq.~\ref{losses} during the training procedure. All the losses converge around 200k optimization epochs.}}}
	\label{losscurve}
\end{figure*}

Fig.~\ref{losscurve} provides plots that describe the changes in all the losses included in Eq.~\ref{losses} of the proposed paper during the training procedure. It is clearly shown that the optimization converges.

\subsubsection{Visual Comparison with the Linear Interpolated Haze Images}

\begin{figure*}
	\centering
	\includegraphics[width=\linewidth]{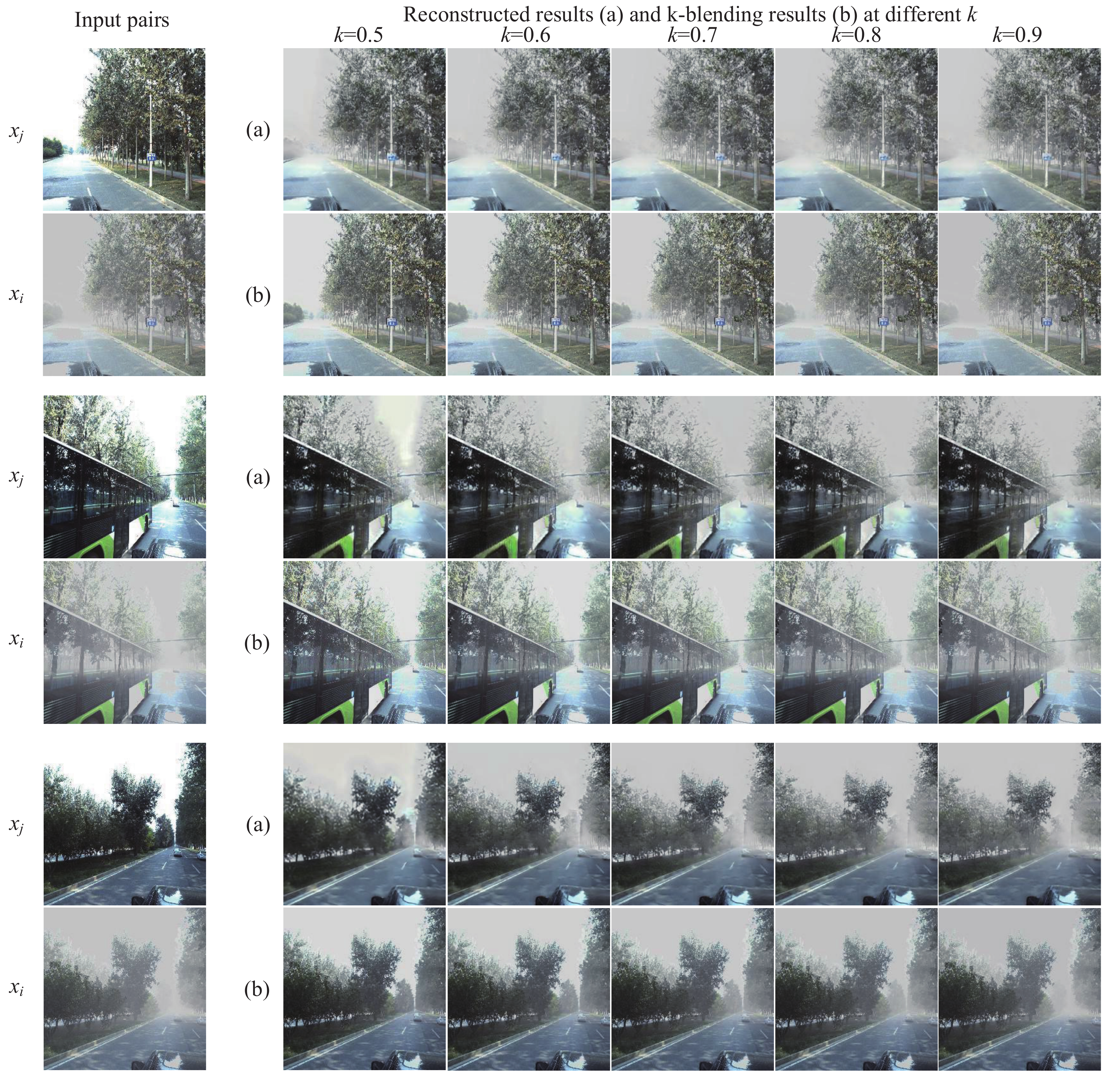}
	\setlength{\abovecaptionskip}{0pt}
	\setlength{\belowcaptionskip}{0pt}
	\caption{Visual comparison between the reconstructed results and the k-blending images of the input pairs shown in left. The images in (a) are the reconstructed images, which are reconstructed from the interpolated style features ($k*s_i+(1-k)*s_j$) and content features $c_j$ (since $c_i$ and $c_j$ describe the same scene, we employ $c_j$ for simplicity). The images in (b) are synthesized by $k*x_i+(1-k)*x_j$.}
	\label{kblending}
\end{figure*}

Fig.~\ref{kblending} shows the results reconstructed from $k*s_i+(1-k)*s_j$ and $c_j$ as well as the k-blending images of the raw image and the corresponding reference haze image, i.e., those synthesized by $k*X_i+(1-k)*X_j$.

The haze-free images $X_j$ and the corresponding reference images $X_i$ are encoded into $s_i, s_j, c_i$ and $c_j$. Then, the style features and input images are interpolated with the parameter $k$. Note that our method does not require depth images, which are required to synthesize the reference haze images in the k-blending method. As shown in Fig.~\ref{kblending}, the haze in the reconstructed results gradually becomes denser as k increases, approaching that of the linear interpolated images and sometimes being even more realistic. For example, caused by the deviation of the depth map, there exists unnatural haze on the surface of the bus in the second group of images in Fig.~\ref{kblending}, while the results synthesized from the interpolated features overcome this problem. Therefore, we state that the disentangled representation learning helps reduce the negative effect caused by errors in the training set when generating multi-density realistic haze images.

\subsubsection{Visualization Results of the t-SNE of Content and Style Features}

\begin{figure*}
	\centering
	\includegraphics[width=\linewidth]{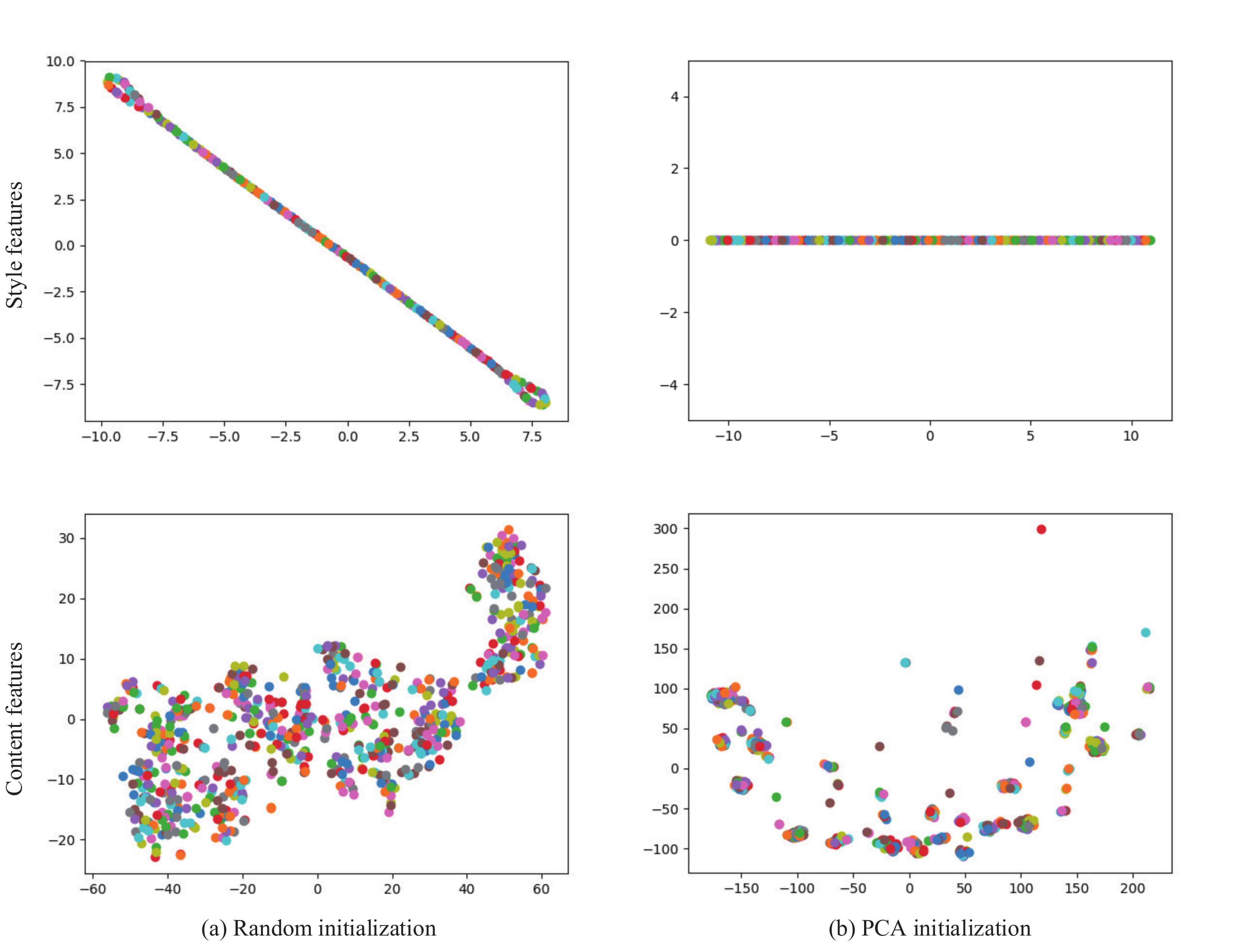}
	\setlength{\abovecaptionskip}{0pt}
	\setlength{\belowcaptionskip}{0pt}
	\caption{The visualization results of the t-SNE of content and style features. The visualization results of the style features are placed at the top, while those of the content features are at the bottom. The columns in (a) and (b) show the results of random initialization and PCA initialization, respectively. {\color{black}{After mapped into 2D space, the style features are aligned in a straight line, which proves that the style features are mapped onto the linear space related to the haze density. In contrast, the content features in both figures are stochastically distributed, which agrees with the fact that the semantic content of each input image is different.}}}
	\label{tsne}
\end{figure*}

We visualized all the style features and content features of the test sets by employing the t-distributed stochastic neighbor embedding (t-SNE) \cite{van2008visualizing}, as shown in Fig.~\ref{tsne}. Specifically, the clear images in the test set are converted to haze images by the proposed method with various haze densities. As in the training process, the density $\alpha$ ranges from 0 to 1. Then, we visualize both the style and content features of these images via the t-SNE, a commonly used visualization method for high-dimensional data. 

If high-dimensional data are linearly distributed, their embeddings in 2D space will form a line. From the results shown in Fig.~\ref{tsne}, we can tell that the style features tend to be aligned with each other in a straight line. Due to the instability of random initialization, the points near both ends of the straight line in the top-left figure are slightly off the line. Therefore, we initialize the embeddings with the more stable PCA. The top-right picture shows that all style features are strictly aligned in a straight line, which proves that the style features are mapped onto the linear space related to the haze density. In contrast, as shown in the second row, the content features in both figures are stochastically distributed, which agrees with the fact that the semantic content of each input image is different. This empirical evidence demonstrates that our method indeed disentangles the content and style as expected.

\subsection{Generalization Analysis}

In this subsection, we give more examples to show the generalization of our model on the in-distribution samples and out-of-distribution samples. Due to the lack of depth images, we are not able to synthesize the corresponding reference images for comparison. We instead compare our synthesized haze images with those synthesized by SOTA image-translation-based dehazing methods \cite{engin2018cycle}\cite{qu2019enhanced}, which can synthesize haze in reverse as well.

\begin{figure*}
	\centering
	\includegraphics[width=\linewidth]{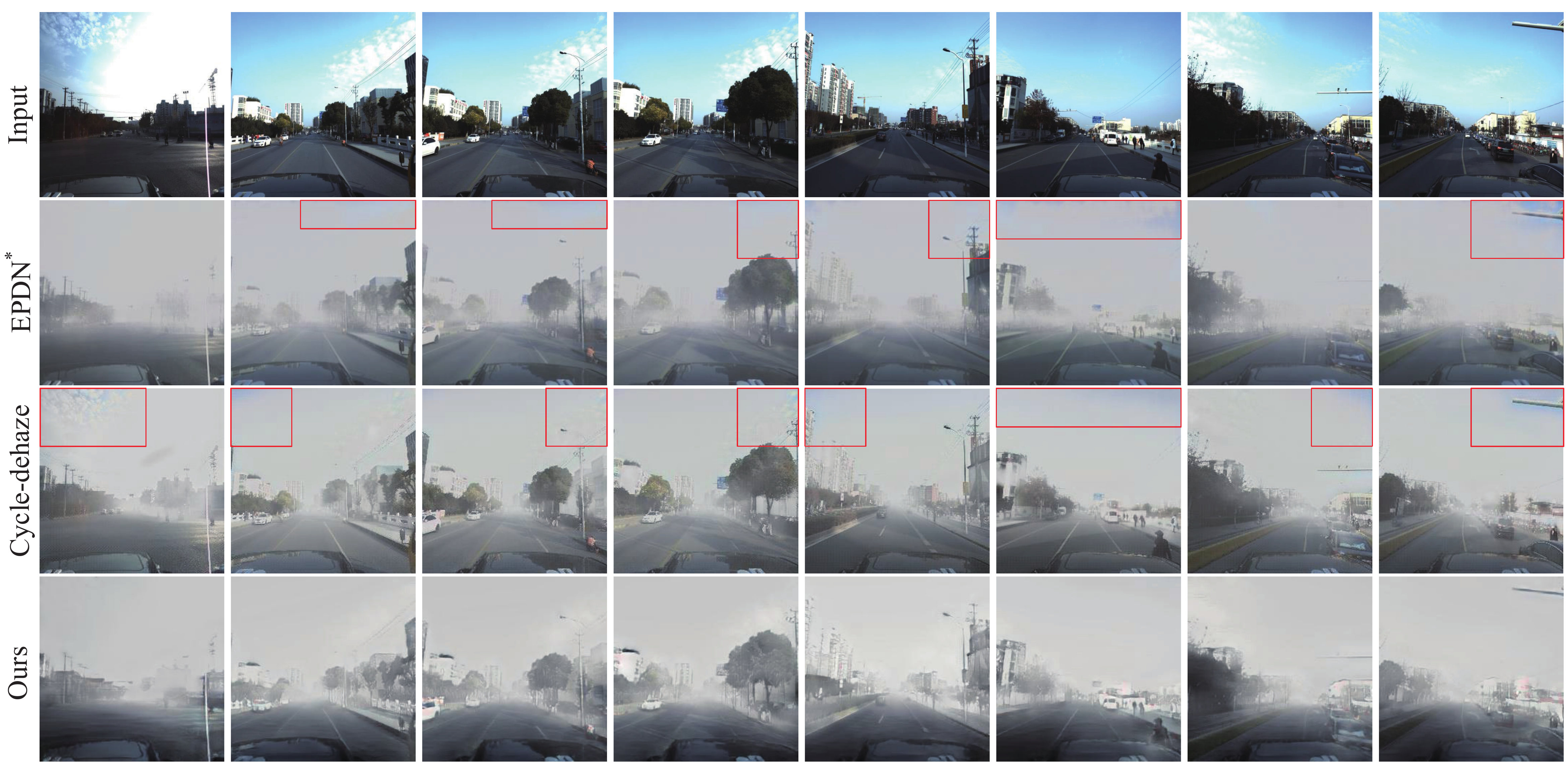}
	\setlength{\abovecaptionskip}{0pt}
	\setlength{\belowcaptionskip}{0pt}
	\caption{{\color{black}{In-distribution generalization of our method, compared with relevant SOTA dehazing models. Test samples are selected from ApolloScape dataset for lane segmentation task. In the results of both the EPDN and Cycle-dehaze, parts of the sky enclosed by the red boxes are still rendered blue, which indicates that mode collapse exists in these two models. Conversely, our model successfully synthesizes complete haze covering the entire scene.}}}
	\label{indis}
\end{figure*}

Fig.~\ref{indis} shows the haze synthesis results on in-distribution samples, which are from the ``Lane Segmentation'' set of ApolloScape. In general, the performances of all models decrease. Compared with the results on the test set (as Fig.~\ref{Fig7} shows), the synthesized haze here is unevenly distributed, with more noise appearing. In the results of both the EPDN and Cycle-dehaze, parts of the sky enclosed by the red boxes are still rendered blue, which indicates that mode collapse exists in these two models. Conversely, our model successfully synthesizes complete haze covering the entire scene. This is because our model learns to render haze that matches the structure of semantic content by disentangling representations. Rather than simply learning pixel mapping relationships as in \cite{engin2018cycle}\cite{qu2019enhanced}, our model shows greater robustness when applied to in-distribution samples.

\begin{figure*}
	\centering
	\includegraphics[width=\linewidth]{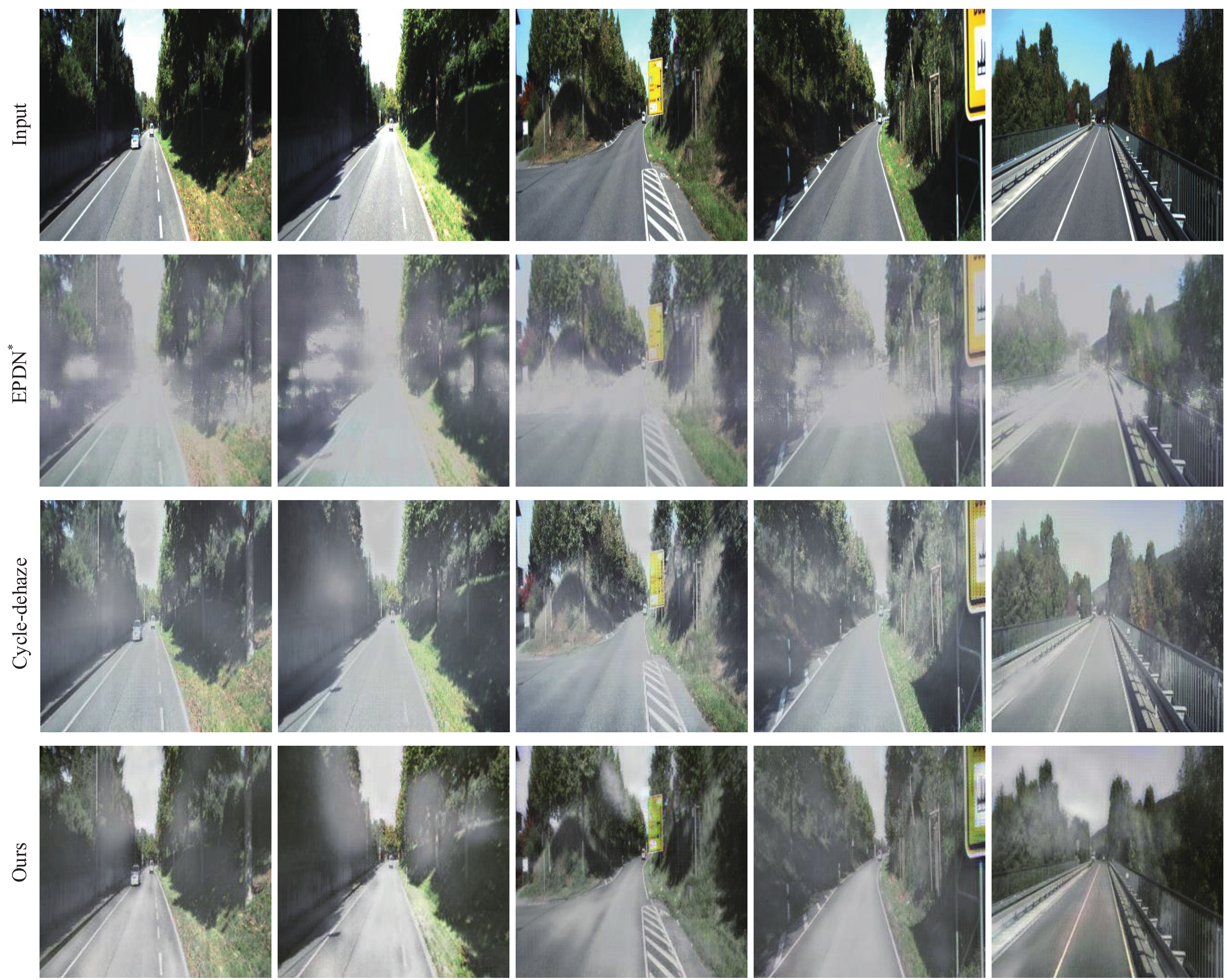}
	\setlength{\abovecaptionskip}{-5pt}
	\setlength{\belowcaptionskip}{-5pt}
	\caption{{\color{black}{Out-of-distribution generalization of our method, compared with relevant SOTA dehazing models. Test samples are selected from the KITTI raw dataset. The performances of all models on the out-of-distribution samples decrease substantially.}}}
	\label{outdis}
\end{figure*}

Fig.~\ref{outdis} shows the haze synthesis results on the KITTI raw dataset \cite{geiger2013vision}. As we can see, the performances of all models on the out-of-distribution samples decrease substantially. The two unsupervised models, Cycle-dehaze and our method, fail to synthesize evenly distributed haze at the vanishing point of the road. The supervised model EPDN synthesizes overly bright and unrealistic haze in the middle of the road, which is inconsistent with the scene structure. Furthermore, the results of both the EPDN and Cycle-dehaze suffer from excessive brightness, which appears as reflection, while our model maintains a more natural tone.

\section{Significance in scenario testing for autonomous driving}
In this section, we set up three experiments to demonstrate that the haze images synthesized by our method provide sufficient difficult scenario data for the testing of the autonomous driving perception module. Specifically, two datasets are organized, one containing only haze-free images and the other additionally containing haze images synthesized from those haze-free images. Then, we explore the distribution change regularity and correlation of the WPS on these two datasets. Furthermore, the reasonableness of generating images with controllable haze density for controlling the difficulty of a scenario is discussed. In this work, based on the framework of the WPS search proposed in our previous work \cite{xu2020worst}, we use YOLOv3 to conduct the WPS search experiments for vehicle detection on the ApolloScape dataset.


\subsection{Regularity of the Distribution Changes of the WPS}
From ApolloScape, we choose the continuous road scenario data as the original dataset, a total of 5773 data points, all of which contain vehicles. Subsequently, we use the original dataset to synthesize the corresponding 5773 haze images with the highest haze density and combine the original image and the synthesized image into a haze dataset, with a total of 11574 images. Notice that the resolution of the synthesized image is inconsistent with the resolution of the original image; thus, we need to downsample the original data to eliminate the influence of image quality on the results of the search experiments. Considering that the resolution is greatly reduced, we filter the labels of the processed data by eliminating the bounding boxes with a length and width less than 5 pixels.

Combining the scenario annotation from the dataset and the image coding from our encoder, we design a scenario search space consisting of 5 scenario-level feature parameters (number of motor vehicles, number of pedestrians, number of non-motor vehicles, number of people, and haze density code) and a 25-dimensional content coding obtained by the content encoder (reducing the dimensionality of the content coding to 25 dimensions through the PCA algorithm). In particular, the haze density code has two values (0 and 1) in the haze dataset but only one value (0) in the original dataset. For the configuration of the framework of the WPS search, we set the number of iterations for each search to 300 and perform 100 search experiments on the original dataset and on the haze dataset.

\begin{table}
	\setlength{\abovecaptionskip}{0pt} 
	\setlength{\belowcaptionskip}{0pt} 
	\caption{F1 score of the test scenario on two datasets, including a comparison of the overall average performance, the average performance of all the worst scenes and the performance of the worst scenes found.}
	\label{table10}
	\centering
	\begin{tabular}{cccc}
		\toprule
		\textbf{Dataset} & \textbf{Avg of all data}& \textbf{Avg of WPS}& \textbf{Worst} \\
		\midrule
		\textbf{Original dataset} & 0.6022 & 0.3626 & 0.0833 \\ 
		\textbf{Haze dataset} & 0.4641 & 0.0906 & 0 \\
		\bottomrule
	\end{tabular}
\end{table} 
\begin{figure}[!t]
	\centering
	\setlength{\abovecaptionskip}{2pt} 
	\setlength{\belowcaptionskip}{0pt} 
	\includegraphics[width=0.5\textwidth]{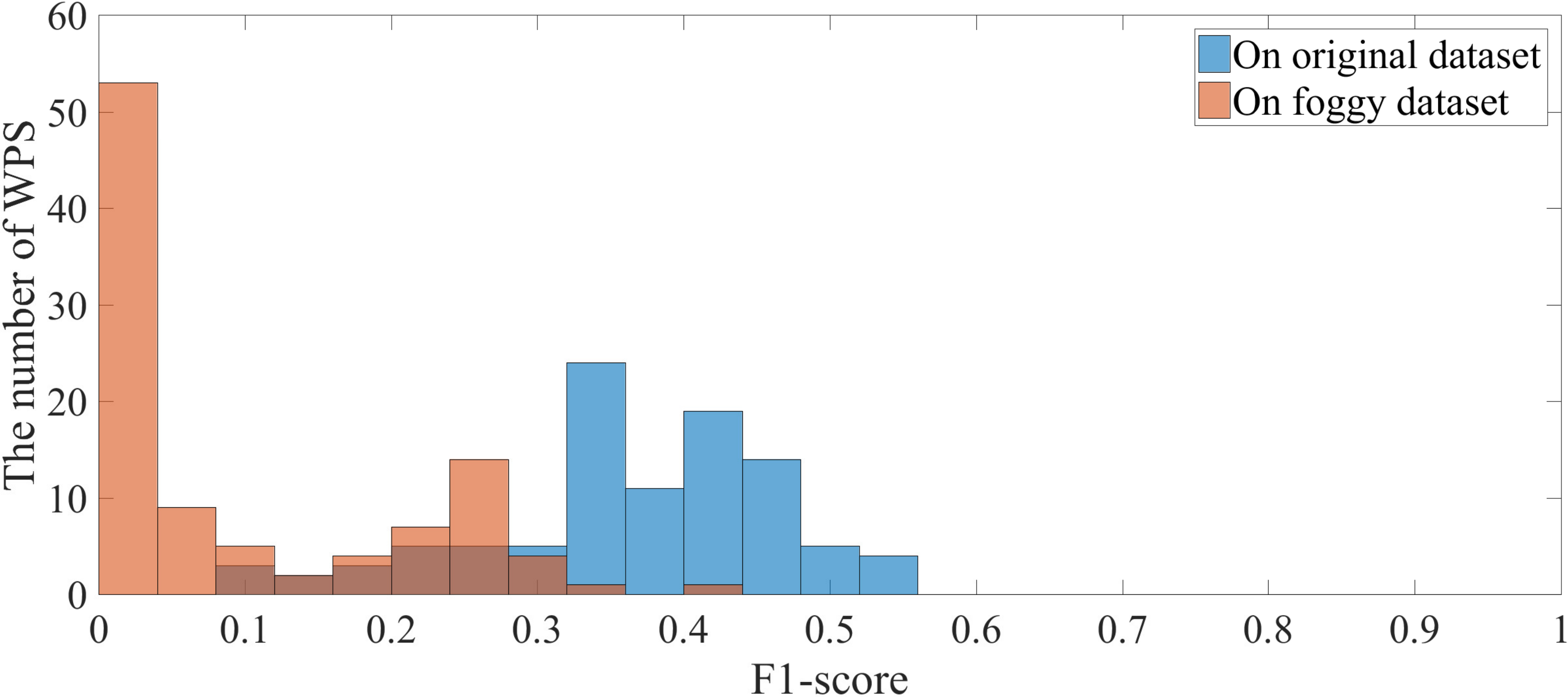}
	\caption{The performance distribution diagram of the WPS searched on the two datasets.}
	\label{fig10}
\end{figure}

Table~\ref{table10} shows a comparison between the overall performance and the performance of the WPS on the two datasets. Intuitively, the overall performance of the perception algorithm on the haze dataset has dropped significantly compared to that of the original dataset; this trend is also manifested in the WPS. This result shows that the addition of synthesized haze images significantly increases the difficulty of the dataset, and this increase is consistent, not only for simple scenarios but also for the WPSs. As shown in Fig.~\ref{fig10}, the change in the performance distribution of the WPS shows the benefit of increasing the upper limit of the difficulty of the worst scenario in the dataset.

\subsection{Correlation of the Distribution Changes of the WPS}
Based on the above experimental search results, we design a transfer experiment to prove the correlation between the WPSs on the two datasets.

\begin{figure}[!t]
	\centering
	\setlength{\abovecaptionskip}{2pt} 
	\setlength{\belowcaptionskip}{0pt} 
	\includegraphics[width=0.5\textwidth]{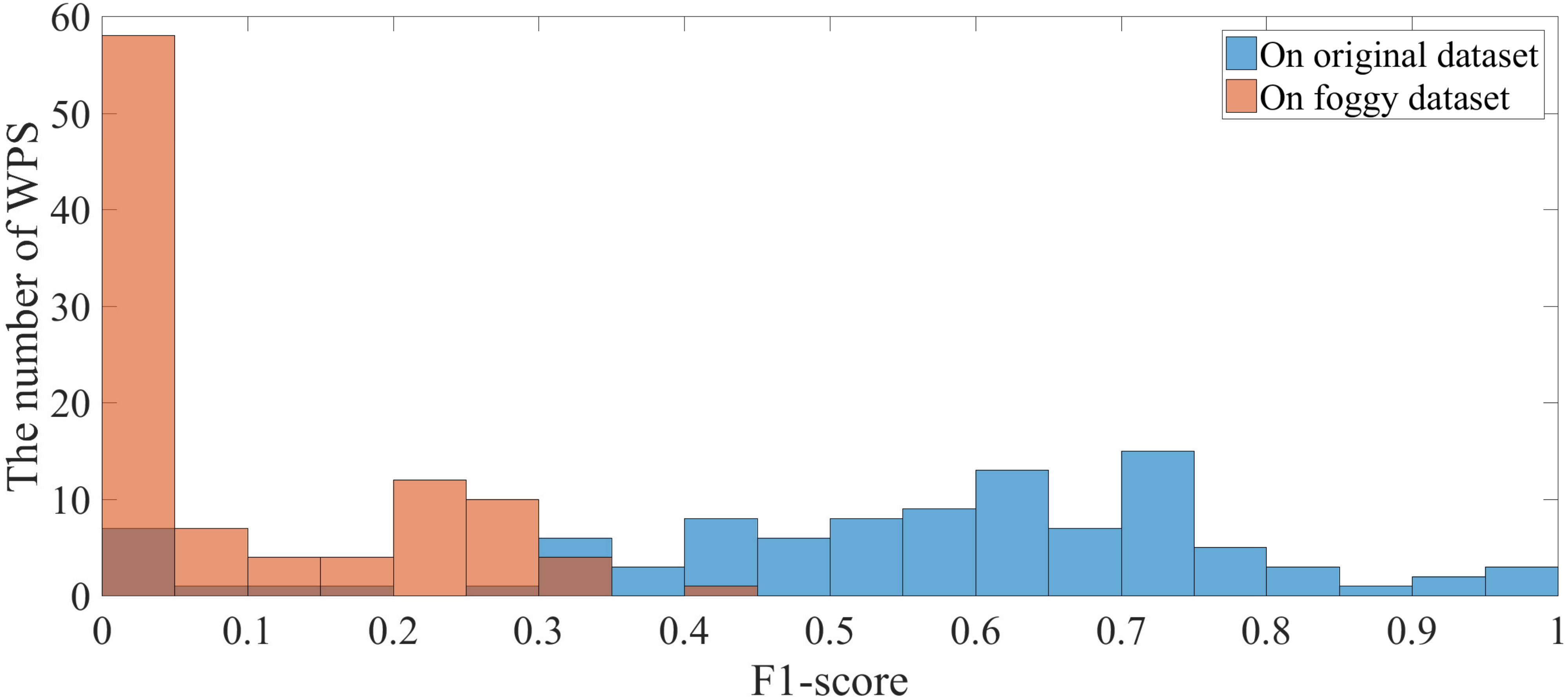}
	\caption{The experimental results of transferring the WPS of the haze dataset to the original dataset.}
	\label{fig11}
\end{figure}

Concretely, since the haze density code in the search space has one value in the original dataset, we can directly transfer the remaining scenario parameters of the WPS from the haze dataset to the original dataset to complete the data matching of the scenario and performance testing. As shown in Fig.~\ref{fig11}, we find that most of the worst scenarios on the haze dataset are simple scenarios in the original dataset, whose performance distribution becomes quite uniform. We preliminarily believe that the result is as follows: the synthesis of haze images degrades the simple scenarios of the original dataset under the control of haze factors. Thus, generating haze images plays an important role in enhancing the richness of difficult scenarios in the dataset.

\begin{figure}[!t]
	\centering
	\setlength{\abovecaptionskip}{0pt} 
	\setlength{\belowcaptionskip}{0pt} 
	\includegraphics[width=0.5\textwidth]{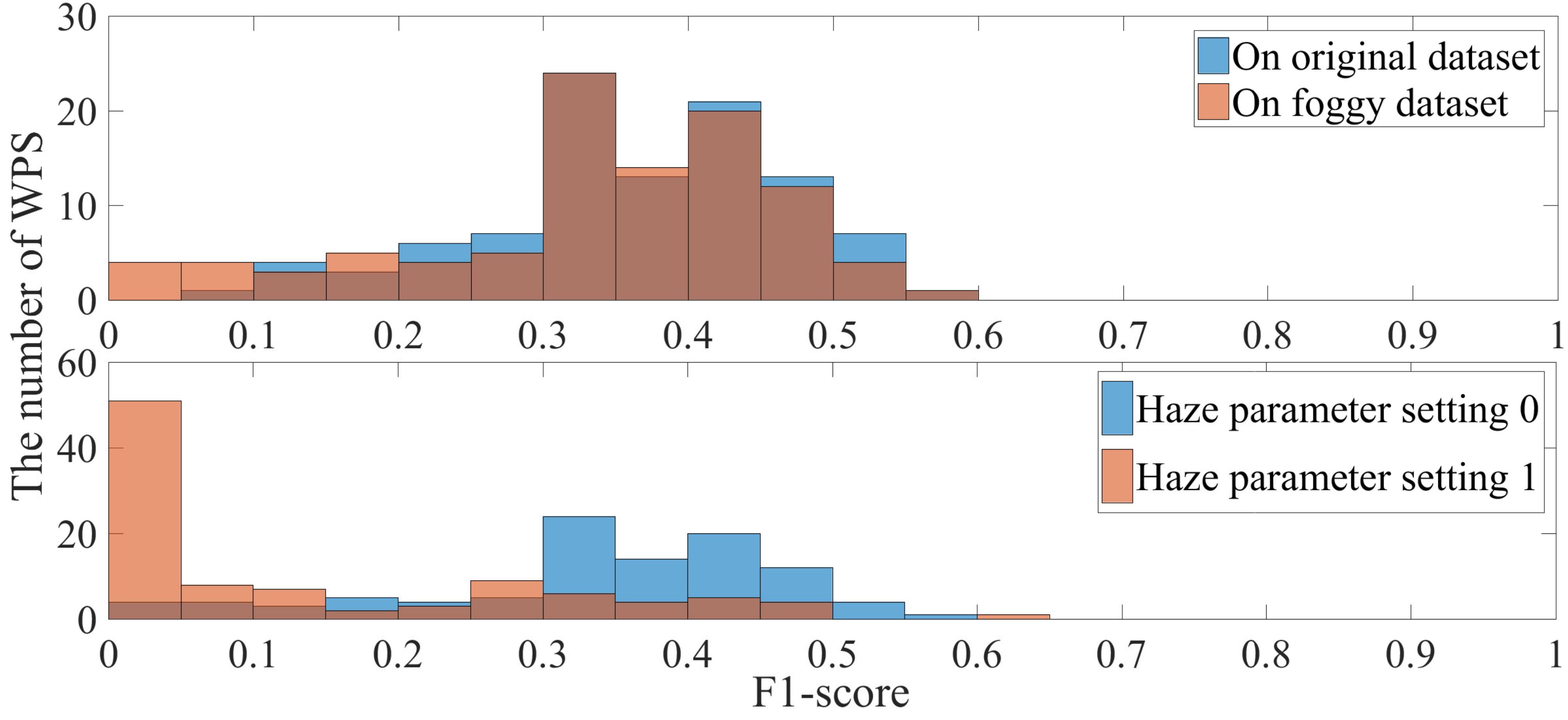}
	\caption{The experimental results of transferring the WPS of the original dataset to the haze dataset. The performance distribution changes of all scenario parameters remain unchanged (the haze density code is always 0), as plotted in the upper figure. The scenario parameters (except the haze density code) remain unchanged (the haze density code is set to 1), and the performance distribution changes are plotted in the bottom figure.}
	\label{fig12}
\end{figure}

In turn, the WPSs on the original dataset are directly transferred to the haze dataset. As expected, since the haze dataset contains the original dataset, Fig.~\ref{fig12} shows that the performance distribution on the two datasets is the same. Based on the WPS of the original dataset, we also implement a comparative experiment in which the haze density code is reversed while the other scenario parameters remain unchanged. Different from the result of direct transfer, the performance degradation becomes obvious, as shown in Fig.~\ref{fig12}. Combining the above two comparisons, we believe that the worst scenarios in the haze dataset include the worst scenarios in the original dataset and that haze has increased the difficulty level of these scenarios.

\subsection{Correlation between the Haze Density and the Difficulty of the Scenario Set}
To explore the effect of the controllable haze density on scenario testing, based on the original dataset, we added various synthesized images with different haze densities, including 0.25, 0.5, 0.625, 0.75, 0.875, and 1. The data processing method is consistent with the above experiment. Then, we obtained a new dataset of varying haze density composed of 40411 images, which is called the multiple haze dataset in this work.

The scenario search space maintains the previous configuration, but the haze density code, one of the scenario-level feature parameters, needs to be converted from (0, 1) to (0, 0.25, 0.5, 0.625, 0.75, 0.875, 1). We set the number of iterations for each search to 300 and perform 125 search experiments on the multiple haze dataset.

\begin{figure}[!t]
	\centering
	\setlength{\abovecaptionskip}{0pt} 
	\setlength{\belowcaptionskip}{0pt} 
	\includegraphics[width=0.5\textwidth]{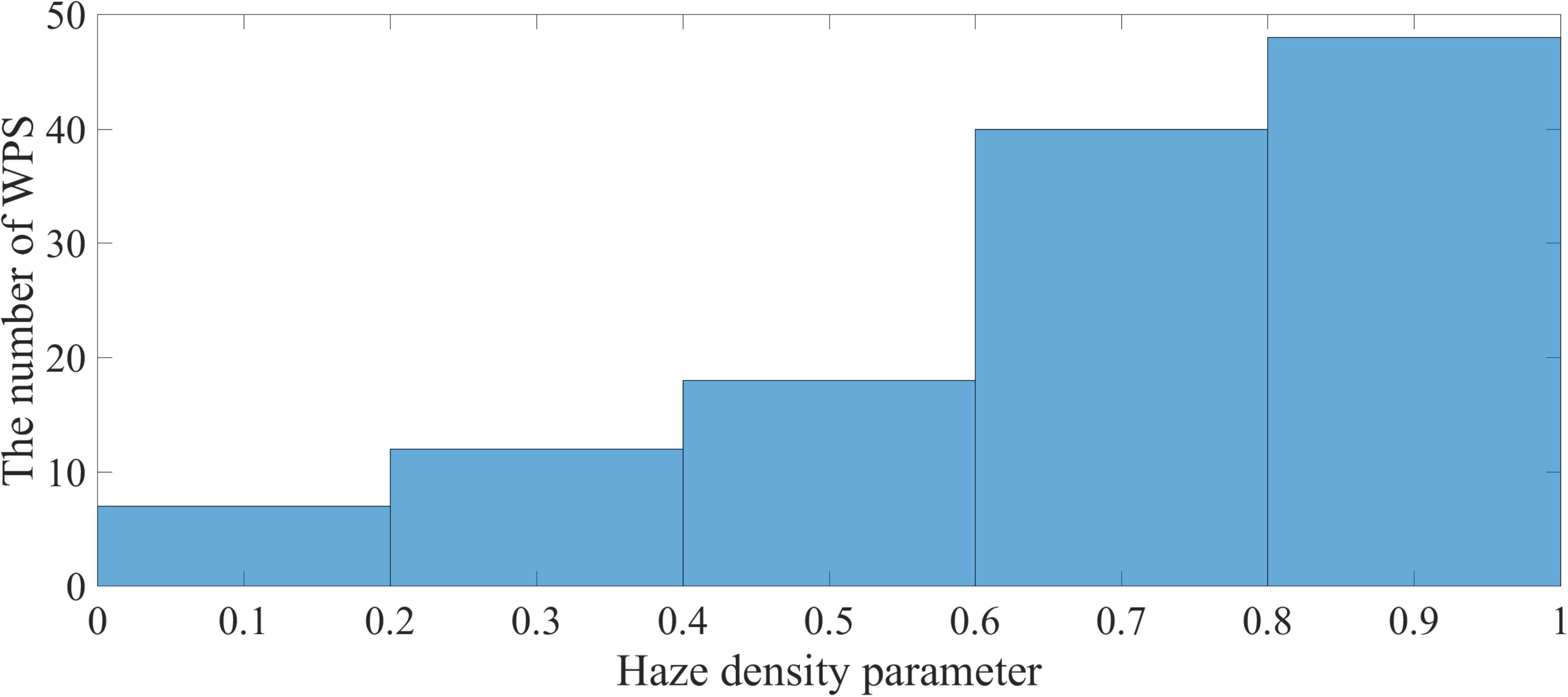}
	\caption{The number of occurrences of different haze density codes in the 125 WPSs searched on the multiple haze dataset.}
	\label{fig14}
\end{figure}

As shown in Fig.~\ref{fig14}, the haze density code of the WPSs obtained by the search are mainly distributed in the high-level range. Therefore, we conclude that the change in haze density may be positively correlated with the degree of influence of scene difficulty.

\begin{table}
	\setlength{\abovecaptionskip}{0pt} 
	\setlength{\belowcaptionskip}{0pt} 
	\caption{Pearson correlation coefficient between scenario-level parameter values and the frequency of WPSs.}
	\label{table11}
	\centering
	\begin{tabular}{cc}
		\toprule
		\textbf{Scenario-level parameter} & \textbf{Pearson correlation coefficient} \\
		\midrule
		\textbf{Vehicle} & -0.4570 \\ 
		\textbf{Person} & 0.2606 \\
		\textbf{Non-motor} & -0.5479 \\
		\textbf{Group} & -0.3134 \\
		\textbf{Haze} & 0.8396 \\
		\bottomrule
	\end{tabular}
\end{table} 

Because the higher the frequency of occurrence of the parameter value in the WPS is, the greater the degree of influence, we use the Pearson coefficient of the scenario parameter value and the frequency of the parameter to measure the correlation between the scenario parameter and the WPS. In Table~\ref{table11}, we calculated the correlation of 5 scenario-level parameters. In previous work, it was found that traffic participants such as the number of vehicles and pedestrians should be strongly correlated with the worst perception, but the correlation here is relatively weak or even negative. We believe that it should be the decorrelation caused by the elimination of the true value of the small target after the data resolution has decreased. The influence of the haze density code shows a strong linear correlation, which provides a reasonable and reliable experimental basis for us to change the haze density code to achieve adaptive difficulty transformation of the scenario testing.

\section{Conclusions}
This paper proposes a density-aware haze image generation method based on disentangled representation learning. By randomly setting the haze density of the synthesized image, we successfully disentangle the haze style feature and the content feature in a self-supervised manner. To learn the mapping from haze-free images to haze images, this paper uses the cycle consistency loss to narrow the implicit space. Moreover, we conduct comparative experiments to analyze the impact of each loss or module in the model on the result. Compared with other methods quantitatively and qualitatively, the results all indicate that our method can synthesize high-quality and high-diversity traffic scene images under the effect of haze. Finally, we demonstrate the obvious negative impact of haze on specific tasks by performing a worst scenario search, which shows the significance in the synthesis of haze scenes for autonomous driving testing.

The direction of future work is threefold:

1. Introduce the proposed method to synthesis tasks involving more adverse weather or lighting conditions, such as rain, snow and nighttime traffic scene images.

2. Further study the theoretical foundations and dynamics of disentangled representation learning for unsupervised image-to-image translation by incorporating the current knowledge on self-supervised learning and contrast learning.

3. Explore haze image synthesis when the distribution of the atmospheric suspended particles is anisotropic, i.e., the synthesis of agglomerate fog images.

\section*{Appendix}

In this appendix, we provide a derivation of Eq.~\ref{equ:sk} based on group theory \cite{scott2012group}. Neural networks are widely utilized to learn the inference process $ f: O \rightarrow Z $ leading from observations to latent space representations. Suppose that we have a group $ G$ of symmetries acting on $ O $ via an action $ \cdot : G \times O \rightarrow O $. If an agent's representation $Z$ is disentangled, the map $f$ should be equivariant between the actions on $O$ and $Z$ \cite{higgins2018towards}. This states that the action ($ \cdot$) should commute with $  f$:


\begin{equation}
\label{definition}
\begin{aligned}
g \cdot f(o) = f(g \cdot o), \forall g \in G, o \in O
\end{aligned}
\end{equation}

Hence, $f$ can be called a G-morphism or an equivariant map ($id_G$ is the identity element of group $G$) \cite{higgins2018towards}:

\begin{equation}
\label{Gmap}
\begin{tikzpicture} 
	\node (A) at (2,1) {$G \times O$};
	\node (B) at (4,1) {$O$};
	\node (C) at (2,-1) {$G \times Z$};
	\node (D) at (4,-1) {$Z$};
	\node (E) at (3.15,1.25) {$\cdot O$};
	\node (F) at (1.3,0) {$id_G \times f$};
	\node (G) at (4.15,0) {$f$};
	\node (H) at (3.15,-0.75) {$\cdot Z$};
	\draw [->] (A.east) -- (B.west) ;
	\draw [->] (A.south) -- (C.north);
	\draw [->] (B.south) -- (D.north);
	\draw [->] (C.east) -- (D.west);
\end{tikzpicture}
\end{equation}


In our case, $O$ refers to the image space, and the latent representations $Z$ are disentangled into the content space $Z_c$ and style space $Z_s$. The inference process $f$ comprises the encoders. Suppose that there are $ o_i $ from the haze image domain and $ o_j $ from the haze-free image domain ($ o_i $ and $ o_j $ are both $256 \times 256$ in size). $ o_i $ and $ o_j $ have style features $ s_i $, $ s_j $ and content features $ c_i $, $c_j$. 

Let the set $\{G_1\}$ denote real numbers ranging from 0 to infinity. Together with the binary operation of multiplication, $(G_1, \times)$ is a group since it satisfies the association, identity and inverse axioms.

In the self-supervised style regression process, we do not care about all values of $Z$ but rather the style space $Z_s$. Thus, we refer to $ f $ as the style encoder. We interpolate style features extracted from two images with randomly selected $k$ ($k$ belongs to $G_1$ and ranges from 0 to 1, as does $(1-k)$) to obtain $ s_k $:



\begin{equation}
\label{adding}
\begin{aligned}
s_k = k f(o_i) + (1-k)f(o_j)
\end{aligned}
\end{equation}

According to Eq.~\ref{definition}, the group action of multiplication on the style space should correspond to the same action on the image space $ O $:

\begin{equation}
\label{22}
\begin{aligned}
k f(o_i) &= f(k o_i) \\
(1-k)f(o_j) &= f((1-k) o_j)\\
\end{aligned}
\end{equation}

Then, consider the group $(G_2, + )$, in which $\{G_2\}$ is the set of 256-dimensional matrices, and the group $(G_3, +)$, in which $\{G_3\}$ is the set of 10-dimensional vectors (with the same length as that of the style features). 
Since $f$ is an equivariant map from $G_2$ to $G_3$, the homomorphism requires that $\forall g_m, g_n \in G_2, f(g_m + g_n) = f(g_m) + f(g_n)$. Thus, the action of adding between style features in Eq.~\ref{adding} should correspond to the action of adding between two images:

\begin{equation}
\label{23}
\begin{aligned}
f(k o_i) + f((1-k) o_j) &= f(k o_i + (1-k)o_j) \\
\end{aligned}
\end{equation}

With Eq.~\ref{adding}, Eq.~\ref{22} and Eq.~\ref{23}, we finally conclude that the action of linear interpolation on style space $Z_s$ will be reflected in image space $O$ as indicated by the following equation:

\begin{equation}
\begin{aligned}
s_k = f(k o_i + (1-k)o_j)
\end{aligned}
\end{equation}

\ifCLASSOPTIONcaptionsoff
  \newpage
\fi



\bibliographystyle{IEEEtran}
\normalem
\bibliography{icme2019template.bib}
%
%

%
\begin{IEEEbiographynophoto}
	{Chi Zhang}(S'14-M'20) received his B.S. and Ph.D. degrees in Control Science and Engineering from Xi'an Jiaotong University, Xi'an, China, in 2011 and 2021, respectively. During 2016 to 2017, he visited the
	Department of Computer Science at Northwestern University, USA, under the supervision of Prof. Ying Wu. He is currently an Assistant Professor with the Institute of Artificial Intelligence and Robotics, Xi'an Jiaotong University, Xi'an, China. His research interests include intelligence testing for autonomous driving systems, computer vision and machine learning.
	
\end{IEEEbiographynophoto}
\begin{IEEEbiographynophoto}
	{Zihang Lin} received a B.E. degree from Xi'an Jiaotong University, Xi’an, China, in 2020. He is currently an M.D. candidate with the College of Artificial Intelligence, Xi’an Jiaotong University, Xi’an, China. His research interests are focused on intelligence evaluations for autonomous vehicles, computer vision and graphics.
\end{IEEEbiographynophoto}

\begin{IEEEbiographynophoto}
	{Liheng Xu} received a B.E. degree from the University of Electronic Science and Technology of China, Chengdu, China, in 2019. He is currently an M.D. candidate with the College of Artificial Intelligence, Xi’an Jiaotong University, Xi’an, China. His research interests are focused on intelligence evaluations for autonomous vehicles, computer vision and graphics.
\end{IEEEbiographynophoto}

\begin{IEEEbiographynophoto}
	{Zongliang Li} received a B.E. degree and an M.E. degree from Xi’an Jiaotong University, Xi’an, China. His research interests are focused on image synthesis for the evaluation of autonomous vehicles, computer vision and graphics.
\end{IEEEbiographynophoto}

\begin{IEEEbiographynophoto}
	{Wei Tang} received his Ph.D. degree in Electrical Engineering from Northwestern University, Evanston, Illinois, USA, in 2019. He received B.E. and M.E. degrees from Beihang University, Beijing, China, in 2012 and 2015, respectively. He is currently an Assistant Professor in the Department of Computer Science at the University of Illinois at Chicago. His research interests include computer vision, pattern recognition and machine learning.
\end{IEEEbiographynophoto}

\begin{IEEEbiographynophoto}
	{Yuehu Liu} received B.E. and M.E. degrees from Xi’an Jiaotong University, Xi’an, China, in 1984 and 1989, respectively, and a Ph.D. degree in Electrical Engineering from Keio University, Tokyo, Japan, in 2000. He is currently a Professor at Xi’an Jiaotong University. His research interests are focused on computer vision, computer graphics, and simulation testing for autonomous vehicles. He is a member of the IEEE and the IEICE.
\end{IEEEbiographynophoto}

\begin{IEEEbiographynophoto}
	{Gaofeng Meng} received a B.S. degree in Applied Mathematics from Northwestern Polytechnical University, Xi’an, China, in 2002, an M.S. degree in Applied Mathematics from Tianjin University, Tianjin, China, in 2005, and a Ph.D. degree in Control Science and Engineering from Xi’an Jiaotong University, Xi’an, China, in 2009. He is currently a Professor with the National Laboratory of Pattern Recognition, Institute of Automation, Chinese Academy of Sciences, Beijing, China. His research interests include document image processing and computer vision. He serves as an Associate Editor for Neurocomputing since 2014. He is a senior member of IEEE.
\end{IEEEbiographynophoto}

\begin{IEEEbiographynophoto}
	{Le Wang} (M’14-SM’20) received the B.S. and Ph.D. degrees in control science and engineering from Xi’an Jiaotong University, Xi’an, China,
	in 2008 and 2014, respectively. From 2013 to 2014, he was a Visiting Ph.D. Student with the Stevens Institute of Technology, Hoboken, New Jersey, USA. From 2016 to 2017, he was a	Visiting Scholar with Northwestern University,
	Evanston, Illinois, USA. He is currently an Associate Professor with the Institute of Artificial	Intelligence and Robotics, Xi’an Jiaotong University, Xi’an, China. His research interests include computer vision,
	pattern recognition, and machine learning. 
\end{IEEEbiographynophoto}

\begin{IEEEbiographynophoto}
	{Li Li} (S'05–M'06–SM'10–F'17)  is currently an Associate Professor with the Department of Automation, Tsinghua University, Beijing, China, where he was involved in artificial intelligence, intelligent control and sensing, intelligent transportation systems, and intelligent vehicles. He serves as an Associate Editor for	the IEEE TRANSACTIONS ON INTELLIGENT TRANSPORTATION SYSTEMS.
\end{IEEEbiographynophoto}





\end{document}